\documentclass[10pt,twocolumn,letterpaper]{article}
\usepackage{cvpr}
\usepackage{times}
\usepackage{epsfig}
\usepackage{graphicx}
\usepackage{amsmath}
\usepackage{amssymb}
\usepackage{subcaption}
\usepackage{float}
\usepackage[utf8]{inputenc}
\usepackage[T1]{fontenc}
\usepackage{capt-of,etoolbox}
\usepackage[export]{adjustbox}
\usepackage{color}
\usepackage{microtype}
\usepackage{cite}

\definecolor{kb_color}{rgb}{.8,0.4,0.0}
\definecolor{sp_color}{rgb}{.4,0.8,0.0}
\definecolor{fl_color}{rgb}{.0,0.4,0.8}
\definecolor{es_color}{rgb}{.0,0.8,0.8}

\newcommand{\fl}[1]{{#1}}

\newcommand{\ignorethis } [1] {}
\def\styimg{reference style image} 

\newcommand{\convlayer}[1]{\textsl{#1}}

\usepackage[hyperindex, pagebackref=true,breaklinks=true,letterpaper=true,colorlinks,bookmarks=false]{hyperref}

 \cvprfinalcopy 

\def\cvprPaperID{2058} 

\title{Deep Photo Style Transfer}
\author{Fujun Luan\\
Cornell University\\
{\tt\small fujun@cs.cornell.edu}
\and
Sylvain Paris\\
Adobe \\
{\tt\small sparis@adobe.com}
\and
Eli Shechtman\\
Adobe \\
{\tt\small elishe@adobe.com}
\and
Kavita Bala\\
Cornell University \\
{\tt\small kb@cs.cornell.edu}
}

\begin{document}

\twocolumn[{%
\renewcommand\twocolumn[1][]{#1}%
\maketitle
\begin{center}
    \centering
    \includegraphics[width=1.0\textwidth]{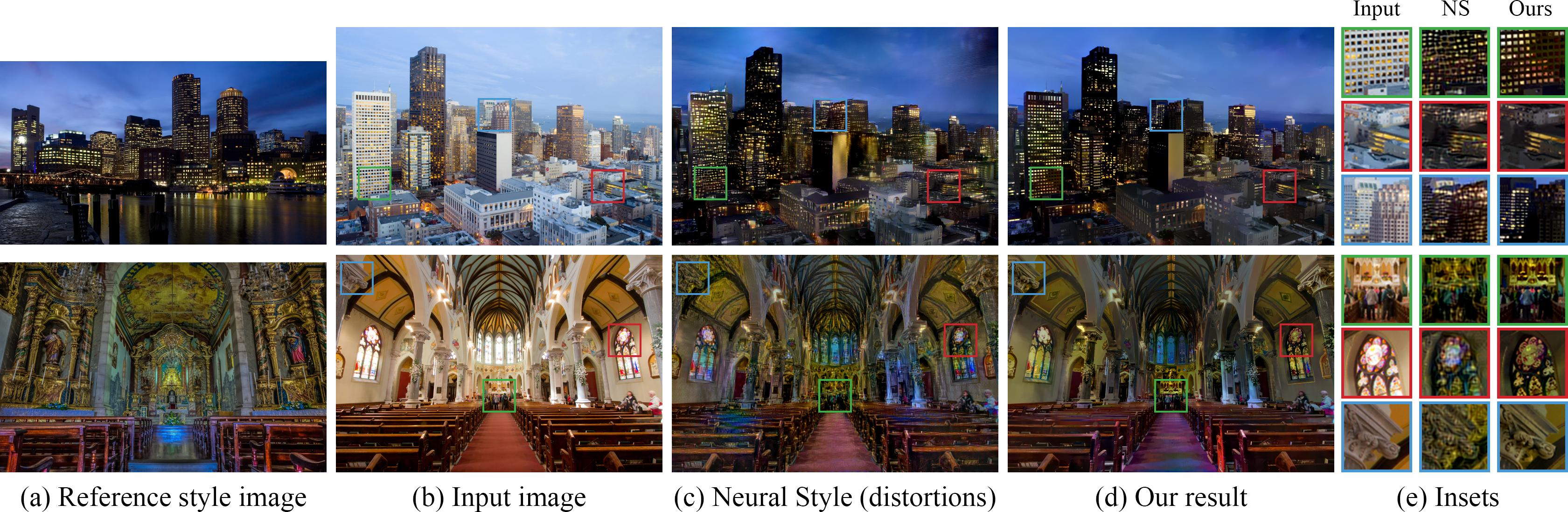}
    \captionof{figure}{Given a reference style image (a) and an input
      image (b), we seek to create an output image of the same scene
      as the input, but with the style of the reference 
      image. The Neural Style
      algorithm~\cite{gatys2016image}~(c) successfully transfers 
      colors, but also introduces distortions that make the output
      look like a painting, which is undesirable in the context of
      photo style transfer. In comparison, our
      result (d) transfers the color of the reference style image
      equally well while preserving the photorealism of the output. On
      the right (e), we show 3 insets of (b),
      (c), and (d) (in that order). Zoom in to compare results. \label{fig:teaser} }
\end{center}%
}]


\begin{abstract}
  This paper introduces a deep-learning approach to photographic style transfer
	that handles a large variety of image content while faithfully
	transferring the reference style. 
	Our approach builds upon the recent work on painterly transfer that
	separates style from the content of an image by considering different
	layers of a neural network. However, as is, this approach is not
	suitable for photorealistic style transfer.  Even when both the input
	and reference images are photographs, the output still exhibits
	distortions reminiscent of a painting. Our contribution is to constrain
	the transformation from the input to the output to be locally affine in
	colorspace, and to express this constraint as a custom fully differentiable energy term.
	We show that this approach successfully suppresses distortion and yields satisfying photorealistic
	style transfers in a broad variety of scenarios, including transfer of
	the time of day, weather, season, and artistic edits.
\end{abstract}

\section{Introduction}
Photographic style transfer is a long-standing problem that seeks to
transfer the style of a reference style photo onto another input
picture. For instance, by appropriately choosing the reference style
photo, one can make the input picture look like it has been taken
under a different illumination, time of day, or weather, or that it
has been artistically retouched with a different intent. So far,
existing techniques are either limited in the diversity of scenes or
transfers that they can handle or in the faithfulness of the stylistic
match they achieve.
In this paper, we introduce a deep-learning approach to photographic
style transfer that is at the same time broad \emph{and} faithful,
i.e., it handles a large variety of image content while accurately
transferring the reference style. Our approach builds upon the recent work on Neural Style transfer by Gatys et al.~\cite{gatys2016image}. However, as shown in Figure~\ref{fig:teaser}, even when the input and reference style images are
photographs, the output still looks like a painting, e.g., straight
edges become wiggly and regular textures wavy.  One of our
contributions is to remove these painting-like effects by preventing
spatial distortion and constraining the transfer operation to happen
only in color space. We achieve this goal with a transformation model
that is locally affine in colorspace, which we express as a custom
{fully differentiable energy term} inspired by the Matting
Laplacian~\cite{levin2008closed}. We show that this approach
successfully suppresses distortion while having a minimal impact
on the transfer faithfulness. Our other key contribution is a solution
to the challenge posed by the difference in content between the input
and reference images, which could result in undesirable transfers between unrelated content.
For example, consider an image with less sky visible in the input image; a
transfer that ignores the difference in context between style and input may
cause the style of the sky to ``spill over'' the rest of the picture. We show
how to address this issue using semantic segmentation~\cite{chen2016deeplab}
of the input and reference images.
We demonstrate the effectiveness of our approach with satisfying photorealistic
style transfers for a broad variety of scenarios including transfer of
the time of day, weather, season, and artistic edits.

\subsection{Challenges and Contributions}

From a practical perspective, our contribution is an effective
algorithm for photographic style transfer suitable for many
applications such as altering the time of day or weather of a picture,
or transferring artistic edits from a photo to another. To achieve this result, we had to address two fundamental challenges.

\paragraph{Structure preservation.}

There is an inherent tension in our objectives. On the one hand, we
aim to achieve very local drastic effects, e.g., to turn on the lights on
individual skyscraper windows (Fig.~\ref{fig:teaser}). On the other
hand, these effects should not distort edges and regular patterns, e.g., so
that the windows remain aligned on a grid. Formally, we seek a transformation
that can strongly affect image colors while having no geometric effect,
i.e., nothing moves or distorts. Reinhard et al.~\cite{reinhard2001color}
originally addressed this challenge with a global color transform. However, by
definition, such a transform cannot model spatially varying effects and
thus is limited in its ability to match the
desired style. More expressivity requires spatially varying effects,
further adding to the challenge of preventing spatial distortion. A
few techniques exist for specific
scenarios~\cite{shih2013data,laffont2014} but the general case remains
unaddressed. Our work directly takes on this challenge and provides a
first solution to restricting the solution space to
photorealistic images, thereby touching on the fundamental task of
differentiating photos from paintings.

\paragraph{Semantic accuracy and transfer faithfulness.}

The complexity of real-world scenes raises another challenge: the
transfer should respect the semantics of the scene. For instance, in a
cityscape, the appearance of buildings should be matched to buildings,
and sky to sky; it is not acceptable to make the sky look like a
building. One plausible approach is to match each input neural patch
with the most similar patch in the style image to minimize the chances
of an inaccurate transfer. This strategy is essentially the one
employed by the CNNMRF method~\cite{li2016combining}. While plausible,
we find that it often leads to results where many input patches
get paired with the same style patch, and/or that entire regions of the
style image are ignored, which generates outputs that poorly match the
desired style.

\begin{figure*}[htp]
\centering
\includegraphics[width=1.0\linewidth]{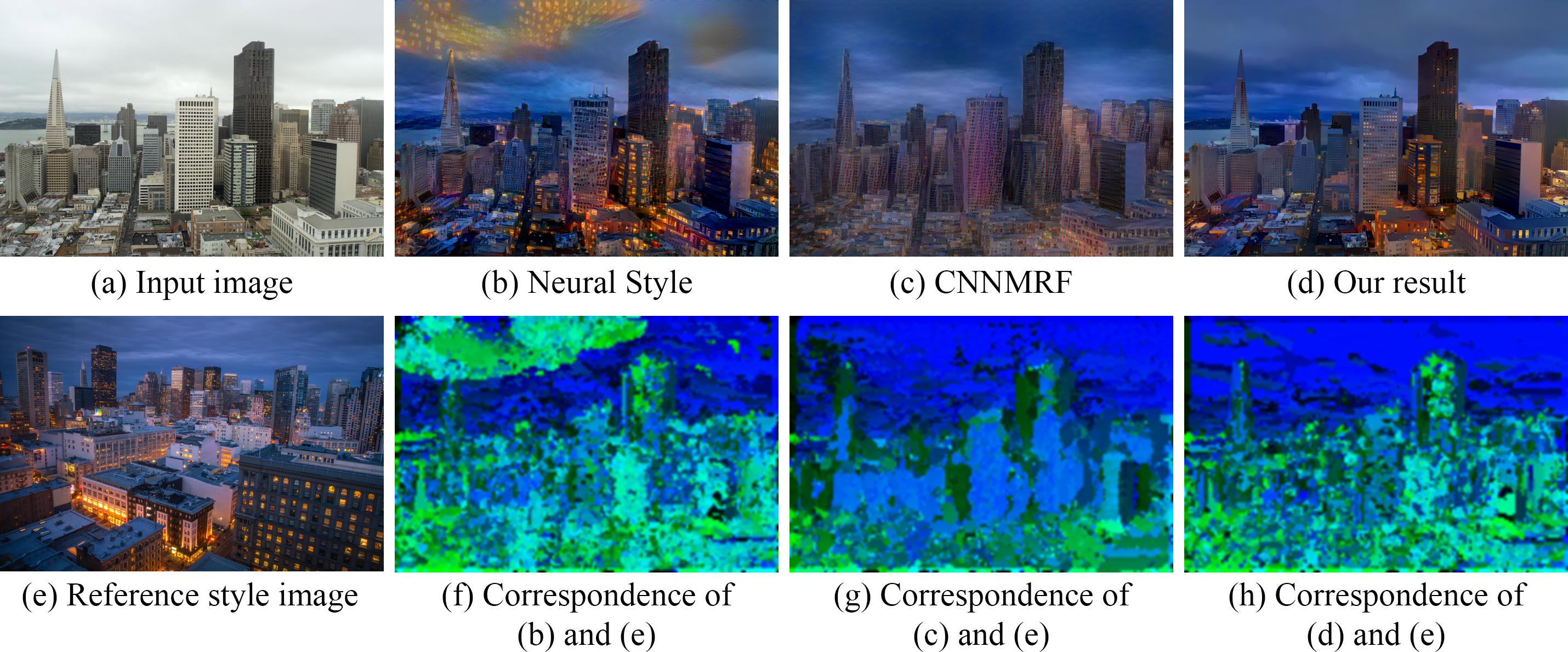}
\caption{Given an input image (a) and a reference style image (e), the results
	(b) of Gatys et al.~\cite{gatys2016image} (Neural Style) and (c) of Li
	et al.~\cite{li2016combining} (CNNMRF) present artifacts due to strong
	distortions when compared to (d) our result.  
	In (f,g,h), we compute the correspondence between the output and the reference
	style where, for each pixel, we encode the $XY$ coordinates of
	the nearest patch in the reference style image as a color with $(R,G,B)
	= (0,\,255\times Y/\mathrm{height},\,255\times X/\mathrm{width})$.
	The nearest neural patch is found using the L2 norm on
	the neural responses of the VGG-19 \convlayer{conv3\_1} layer,
	similarly to CNNMRF.
	Neural Style computes global statistics of the reference style image
	which tends to produce texture mismatches as shown in the
	correspondence (f), e.g., parts of the sky in the output image
	map to the buildings from the reference style image. CNNMRF computes a
	nearest-neighbor search of the reference style image which tends to
	have many-to-one mappings as shown in the correspondence (g), e.g., 
	see the buildings. In comparison, our result (d) prevents
	distortions and matches the texture correctly as shown in the
	correspondence (h).} \label{fig:corr}
\end{figure*}

One solution to this problem is to transfer the complete
``style distribution'' of the reference style photo as captured by the Gram
matrix of the neural responses~\cite{gatys2016image}. This approach
successfully prevents any region from being ignored.  However, there may be some
scene elements more (or less) represented in the
input than in the reference image. In such cases, the style of the large
elements in the reference style image ``spills over'' into mismatching elements
of the input image, generating artifacts like building texture in the
sky. A contribution of our work is to incorporate a semantic labeling
of the input and style images into the transfer procedure so that the transfer
happens between semantically equivalent subregions and within each of them, the
mapping is close to uniform. As we shall see, this algorithm preserves the
richness of the desired style and prevents spillovers. These issues are demonstrated in Figure~\ref{fig:corr}.


\subsection{Related Work}
Global style transfer algorithms process an image by applying a
spatially-invariant transfer function. These methods are effective and
can handle simple styles like global color shifts (e.g., sepia) and
tone curves (e.g., high or low contrast). For instance, Reinhard et
al.~\cite{reinhard2001color} match the means and standard deviations
between the input and \styimg\ after converting them into a decorrelated
color space. Piti\'e et al.~\cite{pitie2005n} describe an algorithm to
transfer the full 3D color histogram using a series of 1D
histograms. As we shall see in the result section, these methods are
limited in their ability to match sophisticated styles.

Local style transfer algorithms based on spatial color mappings are
more expressive and can handle a broad class of applications such as
time-of-day hallucination~\cite{shih2013data,gardner15}, transfer of
artistic edits~\cite{bae2006two,sunkavalli2010multi,shih2014style},
weather and season change~\cite{laffont2014,gardner15}, and painterly
stylization~\cite{hertzmann2001image,gatys2016image,li2016combining,selim2016painting}. Our
work is most directly related to the line of work initiated by Gatys
et al.~\cite{gatys2016image} that employs the feature maps of
discriminatively trained deep convolutional neural networks such as
VGG-19~\cite{simonyan2014very} to achieve groundbreaking performance
for painterly style transfer~\cite{li2016combining,
  selim2016painting}. The main difference with these techniques is
that our work aims for photorealistic transfer, which, as we
previously discussed, introduces a challenging tension between local
changes and large-scale consistency. In that respect, our algorithm is
related to the techniques that operate in the photo realm~\cite{bae2006two,sunkavalli2010multi,shih2014style,shih2013data,gardner15,laffont2014}. But
unlike these techniques that are dedicated to a specific scenario, our
approach is generic and can handle a broader diversity of style
images.


\section{Method}

Our algorithm takes two images: an \emph{input} image which is usually
an ordinary photograph and a stylized and retouched reference image,
the \emph{\styimg}. We seek to transfer the style of the reference to
the input while keeping the result photorealistic. Our approach
augments the Neural Style algorithm~\cite{gatys2016image} by
introducing two core ideas.
\begin{itemize}
\item We propose a photorealism regularization term in the objective function during the optimization, constraining the reconstructed image to be represented by locally affine color transformations of the input to prevent distortions.
\item We introduce an optional guidance to the style transfer process based on semantic segmentation of the inputs (similar to~\cite{neural_doodle}) to avoid the content-mismatch problem, which greatly improves the photorealism of the results.
\end{itemize}

\paragraph{Background.} For completeness, we summarize the
Neural Style algorithm by Gatys et al.~\cite{gatys2016image} that
transfers the reference style image $S$ onto the input image $I$ to produce an
output image $O$ by minimizing the objective function:
\begin{subequations}
\begin{align}
\label{eq:neural_style}
\mathcal{L}_\text{total} &= \sum_{\ell=1}^{L} \alpha_\ell \mathcal{L}_c ^ \ell + \Gamma \sum_{\ell=1}^{L}  \beta_\ell \mathcal{L}_s ^ \ell
\\[10pt]
\label{eq:neural_style_c}
\text{with: } \mathcal{L}_c ^ \ell &\textstyle = \frac{1}{2N_\ell D_\ell} \sum_{ij} (F_\ell[O] - F_\ell[I])_{ij}^2
\\
\label{eq:neural_style_s}
\mathcal{L}_s ^ \ell &\textstyle = \frac{1}{2N_\ell^2} \sum_{ij}(G_\ell[O] - G_\ell[S])_{ij}^2
\end{align}
\end{subequations}
where $L$ is the total number of convolutional layers and $\ell$ indicates the $\ell$-th convolutional layer of the deep convolutional neural network. In each layer, there are $N_\ell$ filters each with a vectorized feature map of size $D_\ell$. $F_\ell[\cdot] \in \mathbb{R}^{N_\ell \times D_\ell}$ is the feature matrix with $(i,j)$ indicating its index and the Gram matrix $G_\ell[\cdot] = F_\ell[\cdot]F_\ell[\cdot]^T \in \mathbb{R}^{N_\ell \times N_\ell}$ is defined as the inner product between the vectorized feature maps. $\alpha_\ell$ and $\beta_\ell$ are the weights to configure layer preferences and $\Gamma$ is a weight that balances the tradeoff between the {\em content} (Eq.~\ref{eq:neural_style_c}) and the {\em style} (Eq.~\ref{eq:neural_style_s}). 

\paragraph{Photorealism regularization.}

We now describe how we regularize this optimization scheme to preserve
the structure of the input image and produce photorealistic
outputs. Our strategy is to express this constraint not on the output
image directly but on the transformation that is applied to the input
image. Characterizing the space of photorealistic images is an
unsolved problem. Our insight is that we
do not need to solve it if we exploit the fact that the input is
already photorealistic. Our strategy is to ensure that we do not lose
this property during the transfer by adding a term to
Equation~\ref{eq:neural_style} that penalizes image distortions. Our
solution is to seek an image transform that is locally affine in color
space, that is, a function such that for each output patch, there is an
affine function that maps the input RGB values onto their output
counterparts. Each patch can have a different affine function, which
allows for spatial variations. To gain some intuition, one can
consider an edge patch. The set of affine combinations of the RGB
channels spans a broad set of variations but the edge itself cannot
move because it is located at the same place in all channels.

Formally, we build upon the Matting Laplacian of Levin et
al.~\cite{levin2008closed} who have shown how to express a grayscale
matte as a locally affine combination of the input RGB channels. They
describe a least-squares penalty function that can be minimized with
a standard linear system represented by a matrix $\mathcal{M}_I$ that
only depends on the input image $I$ (We refer to the original article
for the detailed derivation. {Note that given an input image $I$ with $N$ pixels, $\mathcal{M}_I$ is $N \times N$}).
{We name $V_c[O]$ the vectorized version ($N \times 1$) of the output image $O$ in channel $c$ and
define the
following regularization term that
penalizes outputs that are not well explained by a locally affine transform:
\begin{equation} \label{eq:matting_laplacian}
\mathcal{L}_{m} = \sum_{c=1}^3 V_c[O]^T \mathcal{M}_I V_c[O]
\end{equation}}

\begin{figure*}[htp]
\centering
\subcaptionbox{Input and Style}
  [.166\textwidth]{\includegraphics[width=0.16\linewidth]{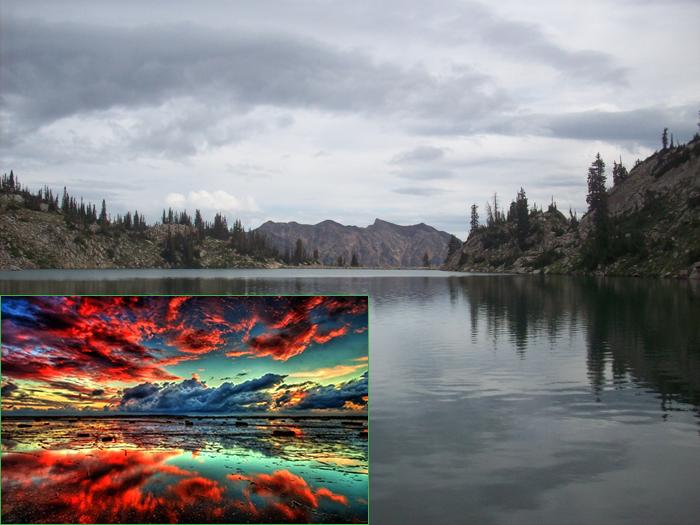}}%
\subcaptionbox{$\lambda=1$}
  [.166\textwidth]{\includegraphics[width=0.16\linewidth]{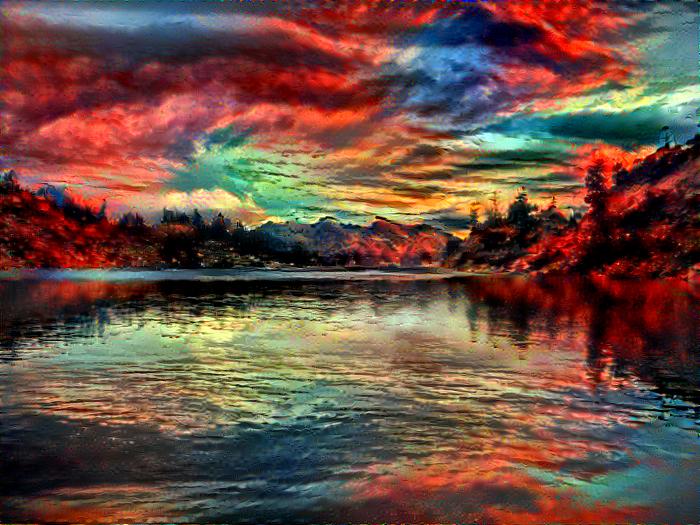}}%
  \subcaptionbox{$\lambda=10^2$}
  [.166\textwidth]{\includegraphics[width=0.16\linewidth]{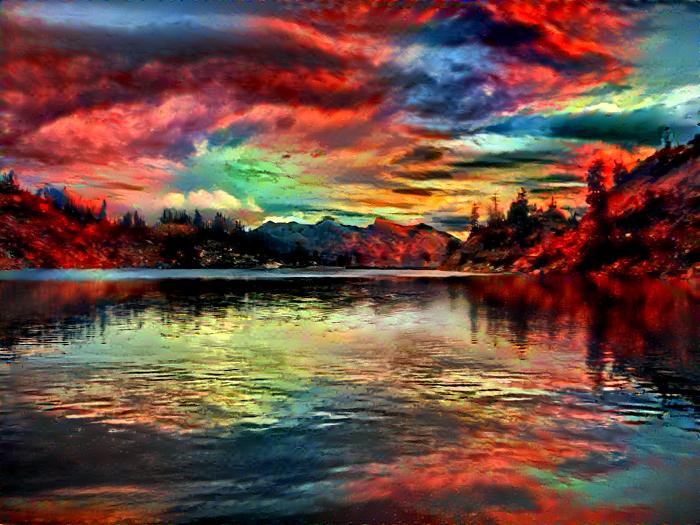}}%
\subcaptionbox{$\lambda=10^4$, our result}
  [.166\textwidth]{\includegraphics[width=0.16\linewidth]{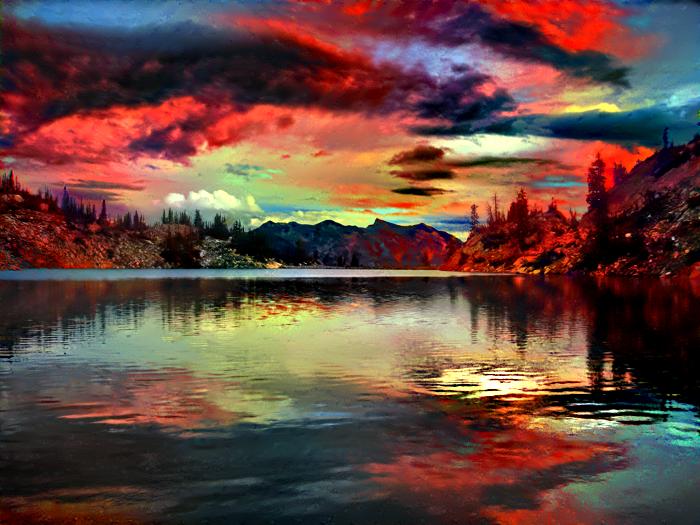}}%
\subcaptionbox{$\lambda=10^6$}
  [.166\textwidth]{\includegraphics[width=0.16\linewidth]{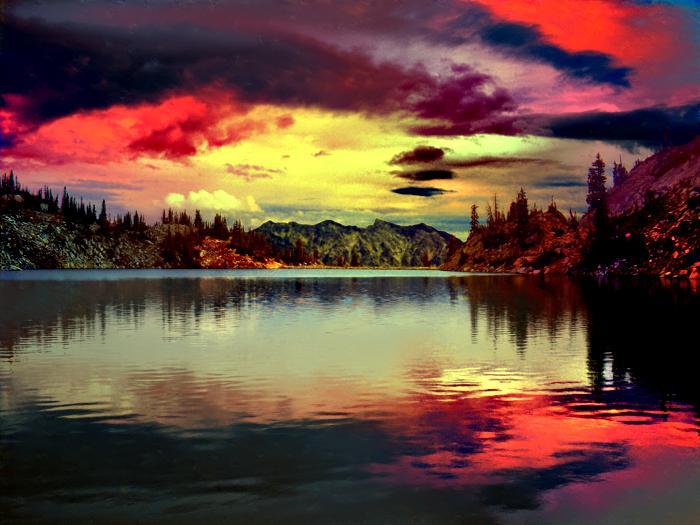}}%
\subcaptionbox{$\lambda=10^8$}
  [.166\textwidth]{\includegraphics[width=0.16\linewidth]{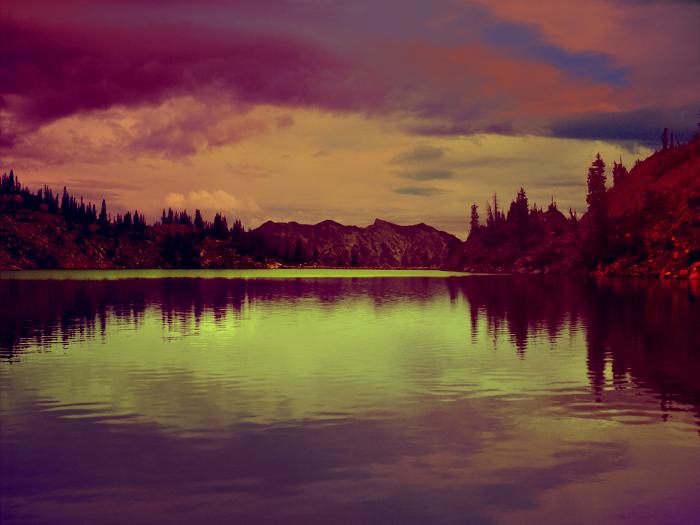}}
	\caption{{Transferring the dramatic appearance of the reference style image ((a)-inset), onto an ordinary flat shot in (a) is challenging. We produce results using our method with different $\lambda$ parameters. A too small $\lambda$ value cannot prevent distortions, and thus the results have a non-photorealistic look in (b,c).
	Conversely, a too large $\lambda$ value suppresses the style to be transferred yielding a half-transferred look in (e,f). We found the best parameter $\lambda=10^4$ to be the sweet spot to produce our result (d) and all the other results in this paper.} }
\label{fig:lambda}
\end{figure*}

{Using this term in a gradient-based solver}
requires us to compute its derivative w.r.t.\ the output image. Since  $\mathcal{M}_I$ is a symmetric matrix, we have: {$\frac{\mathrm{d}\mathcal{L}_{m}}{\mathrm{d}V_c[O]} = 2 \mathcal{M}_I V_c[O].$}

\paragraph{Augmented style loss with semantic segmentation.}

A limitation of the style term (Eq.~\ref{eq:neural_style_s}) is that
the Gram matrix is computed over the entire image. Since a Gram matrix
determines its constituent vectors up to an isometry~\cite{gram}, it
implicitly encodes the exact distribution of neural responses, which
limits its ability to adapt to variations of semantic context and can
cause ``spillovers''. We address this problem with an approach akin to
Neural Doodle~\cite{bae2006two} and a semantic segmentation
method~\cite{chen2016deeplab} to generate image segmentation masks for
the input and reference images for a set of common
labels (sky, buildings, water, etc.). We add the masks to the input image
as additional channels and augment the neural style algorithm by
concatenating the segmentation channels and updating the
style loss as follows:
\begin{subequations}
\begin{align}   \label{eq:neural_style_s_plus}
\mathcal{L}_{s+}^\ell = \sum_{c=1} ^ C \frac{1}{2 N_{\ell,c}^2} \sum_{ij}(G_{\ell,c}[O]  - G_{\ell,c}[S])_{ij}^2
\\
F_{\ell,c}[O] = F_\ell[O] M_{\ell,c}[I] \quad
F_{\ell,c}[S] = F_\ell[S] M_{\ell,c}[S]
\end{align}
\end{subequations}
where $C$ is the number of channels in the semantic segmentation mask,
$M_{\ell,c}[\cdot]$ denotes the channel $c$ of the segmentation mask
in layer $\ell$, and $G_{\ell,c}[\cdot]$ is the Gram matrix
corresponding to $F_{\ell,c}[\cdot]$.  We downsample the masks to
match the feature map spatial size at each layer of the convolutional
neural network.

To avoid ``orphan semantic labels'' that are only present in
the input image, we constrain the input semantic labels to be chosen
among the labels of the reference style image. While this may cause erroneous
labels from a semantic standpoint, the selected labels are in general
equivalent in our context, e.g., ``lake'' and ``sea''. We have also
observed that the segmentation does not need to be pixel accurate
since eventually the output is constrained by our regularization.


\paragraph{Our approach.} We formulate the photorealistic style transfer objective by combining all 3 components together:
\begin{equation} \label{eq:ours}
\mathcal{L}_\text{total} = \sum_{l=1}^{L} \alpha_\ell \mathcal{L}_c ^ \ell + \Gamma \sum_{\ell=1}^{L}  \beta_\ell \mathcal{L}_{s+} ^ \ell + \lambda \mathcal{L}_m
\end{equation}
where $L$ is the total number of convolutional layers and $\ell$ indicates the $\ell$-th convolutional layer of the deep neural network. $\Gamma$ is a weight that controls the style loss. $\alpha_\ell$ and $\beta_\ell$ are the weights to configure layer preferences. $\lambda$ is a weight that controls the photorealism regularization. $\mathcal{L}_c^\ell$ is the content loss (Eq.~\ref{eq:neural_style_c}). $\mathcal{L}_{s+}^\ell$ is the augmented style loss (Eq.~\ref{eq:neural_style_s_plus}). $\mathcal{L}_m$ is the photorealism regularization (Eq.~\ref{eq:matting_laplacian}).


\begin{figure*}[htp]
\begin{subfigure}{.2\textwidth}
	\includegraphics[width=0.95\linewidth, left]{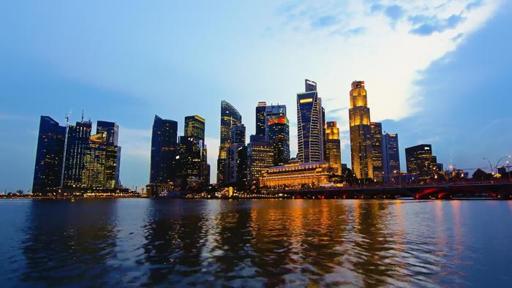}
\end{subfigure}%
\begin{subfigure}{.2\textwidth}
	\includegraphics[width=0.95\linewidth, left]{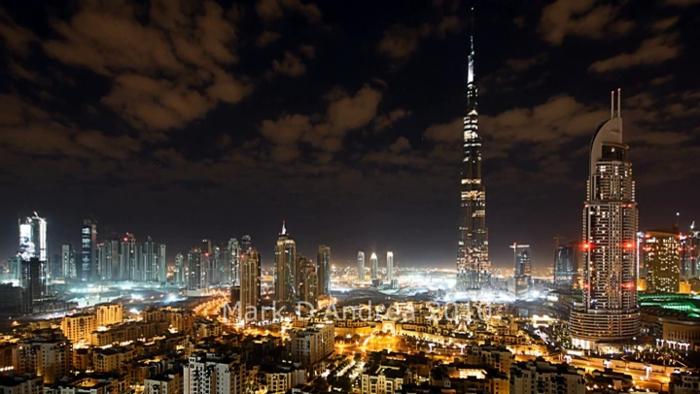}
\end{subfigure}%
\begin{subfigure}{.2\textwidth}
	\includegraphics[width=0.95\linewidth, right]{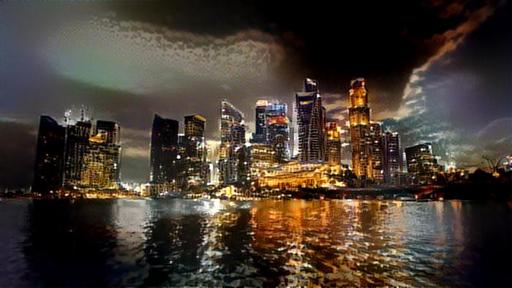}
\end{subfigure}%
\begin{subfigure}{.2\textwidth}
	\includegraphics[width=0.95\linewidth, right]{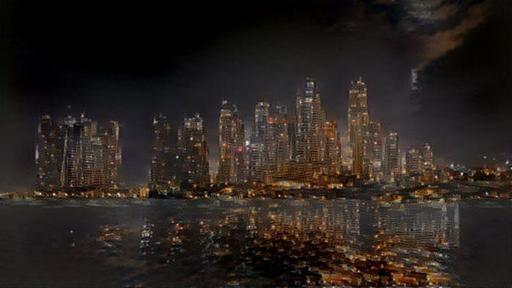}
\end{subfigure}%
\begin{subfigure}{.2\textwidth}
	\includegraphics[width=0.95\linewidth, right]{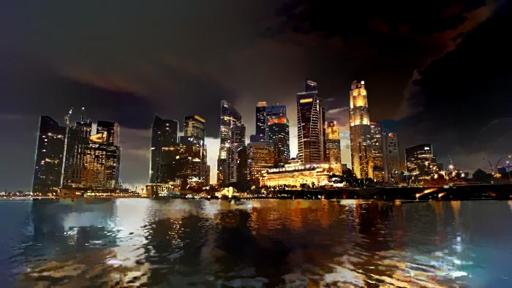}
\end{subfigure}%

\begin{subfigure}{.2\textwidth}
	\includegraphics[width=0.95\linewidth, left]{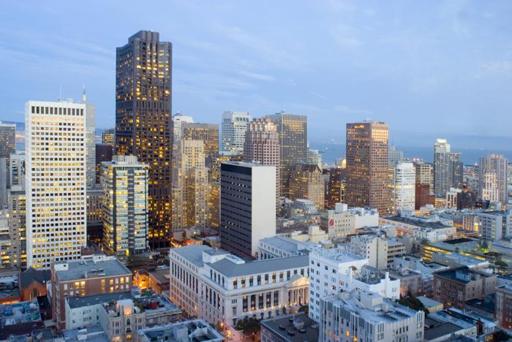}
\end{subfigure}%
\begin{subfigure}{.2\textwidth}
	\includegraphics[width=0.95\linewidth, left]{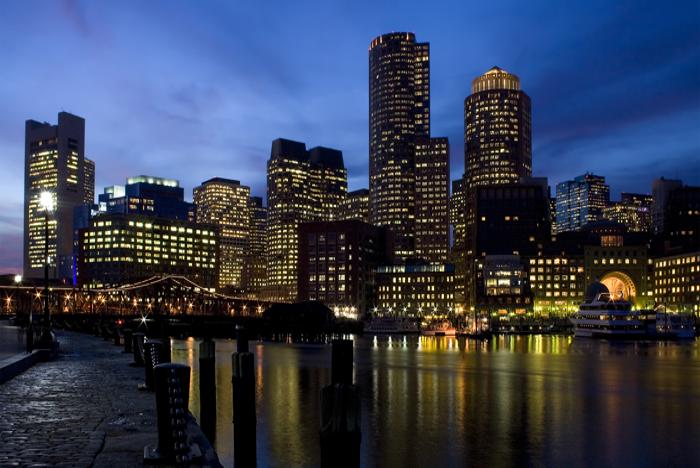}
\end{subfigure}%
\begin{subfigure}{.2\textwidth}
	\includegraphics[width=0.95\linewidth, right]{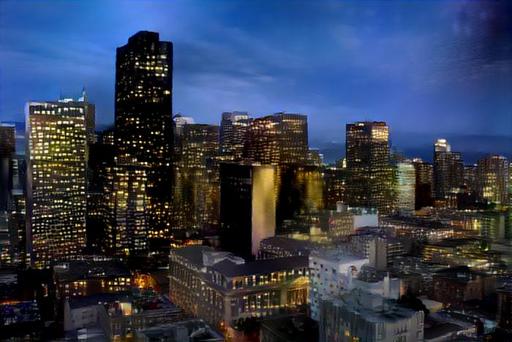}
\end{subfigure}%
\begin{subfigure}{.2\textwidth}
	\includegraphics[width=0.95\linewidth, right]{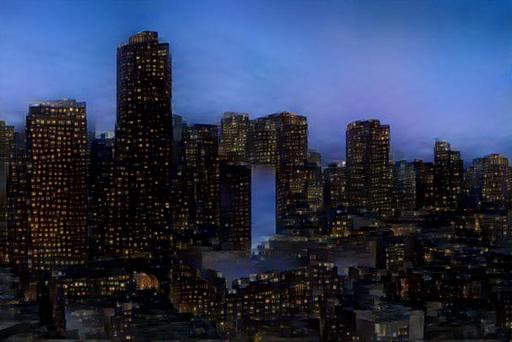}
\end{subfigure}%
\begin{subfigure}{.2\textwidth}
	\includegraphics[width=0.95\linewidth, right]{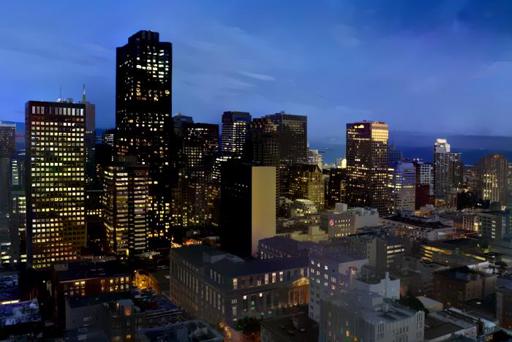}
\end{subfigure}%

\begin{subfigure}{.2\textwidth}
	\includegraphics[width=0.95\linewidth, left]{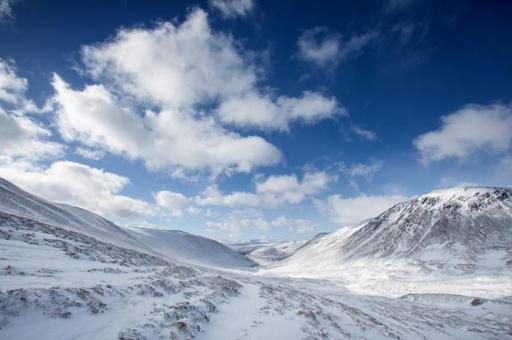}
\end{subfigure}%
\begin{subfigure}{.2\textwidth}
	\includegraphics[width=0.95\linewidth, left]{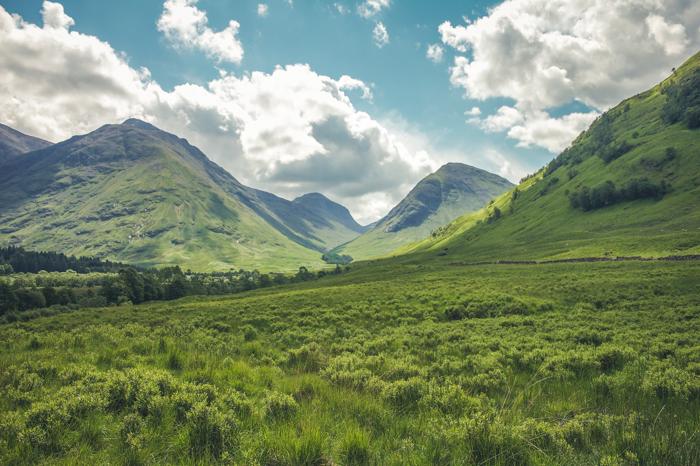}
\end{subfigure}%
\begin{subfigure}{.2\textwidth}
	\includegraphics[width=0.95\linewidth, right]{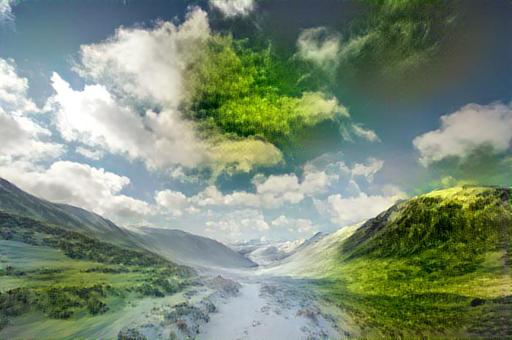}
\end{subfigure}%
\begin{subfigure}{.2\textwidth}
	\includegraphics[width=0.95\linewidth, right]{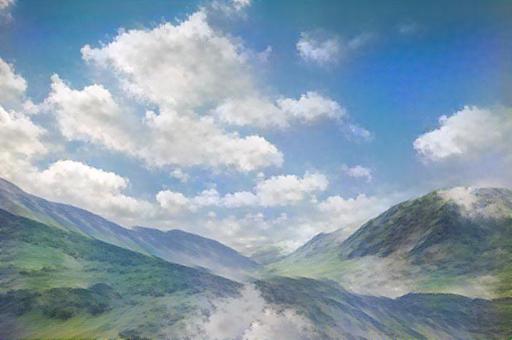}
\end{subfigure}%
\begin{subfigure}{.2\textwidth}
	\includegraphics[width=0.95\linewidth, right]{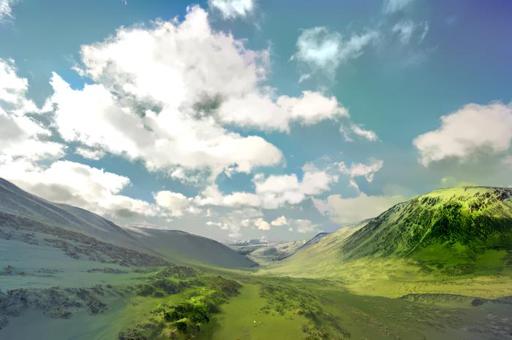}
\end{subfigure}%

\begin{subfigure}{.2\textwidth}
	\includegraphics[width=0.95\linewidth, left]{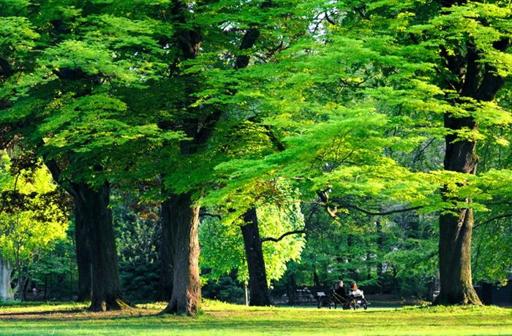}
\end{subfigure}%
\begin{subfigure}{.2\textwidth}
	\includegraphics[width=0.95\linewidth, left]{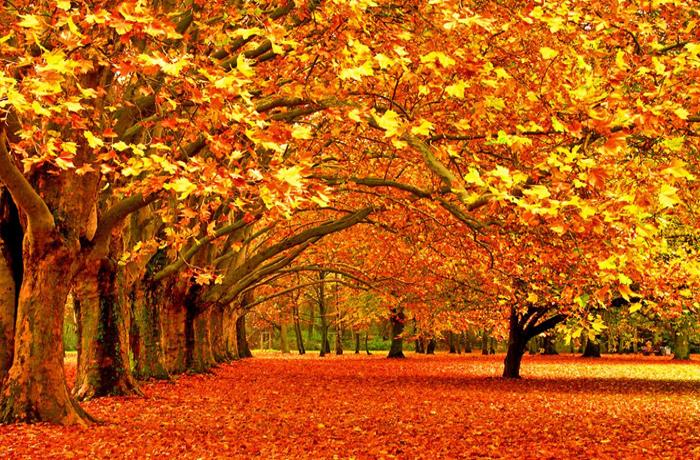}
\end{subfigure}%
\begin{subfigure}{.2\textwidth}
	\includegraphics[width=0.95\linewidth, right]{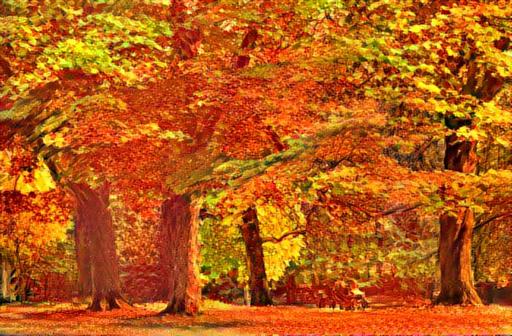}
\end{subfigure}%
\begin{subfigure}{.2\textwidth}
	\includegraphics[width=0.95\linewidth, right]{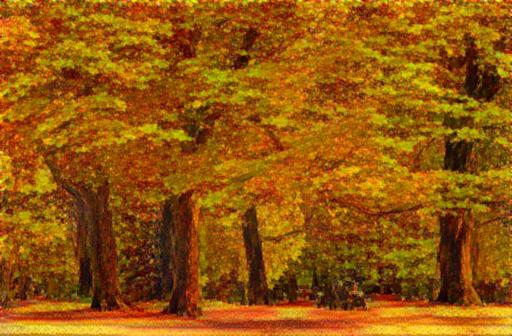}
\end{subfigure}%
\begin{subfigure}{.2\textwidth}
	\includegraphics[width=0.95\linewidth, right]{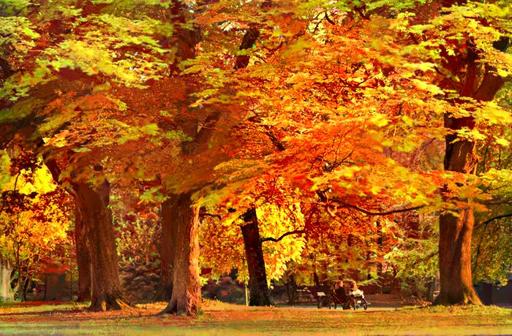}
\end{subfigure}%

\begin{subfigure}{.2\textwidth}
	\includegraphics[width=0.95\linewidth, left]{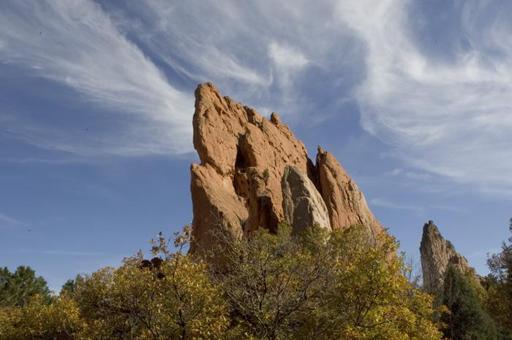}
\end{subfigure}%
\begin{subfigure}{.2\textwidth}
	\includegraphics[width=0.95\linewidth, left]{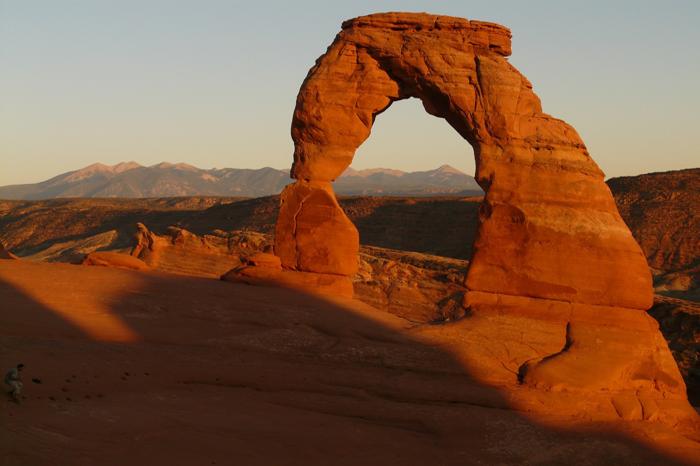}
\end{subfigure}%
\begin{subfigure}{.2\textwidth}
	\includegraphics[width=0.95\linewidth, right]{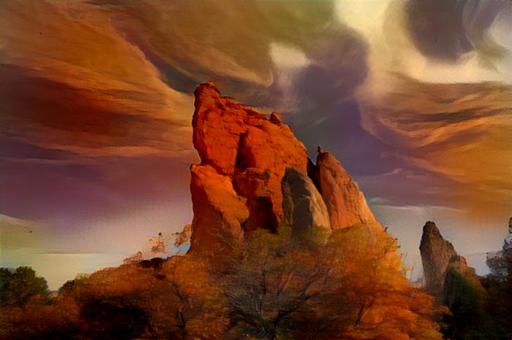}
\end{subfigure}%
\begin{subfigure}{.2\textwidth}
	\includegraphics[width=0.95\linewidth, right]{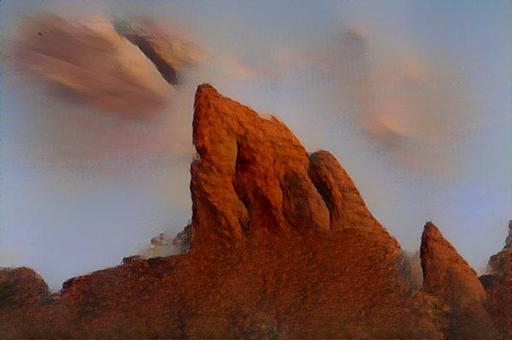}
\end{subfigure}%
\begin{subfigure}{.2\textwidth}
	\includegraphics[width=0.95\linewidth, right]{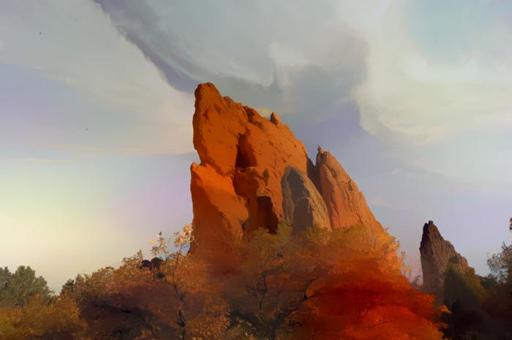}
\end{subfigure}%

\begin{subfigure}{.2\textwidth}
	\includegraphics[width=0.95\linewidth, left]{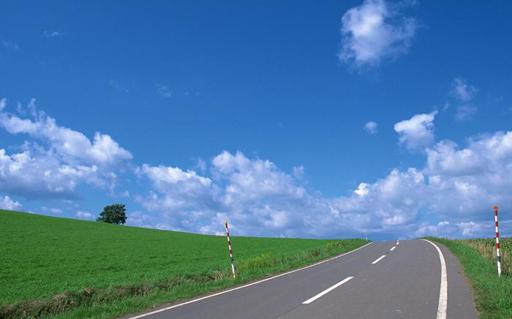}
\end{subfigure}%
\begin{subfigure}{.2\textwidth}
	\includegraphics[width=0.95\linewidth, left]{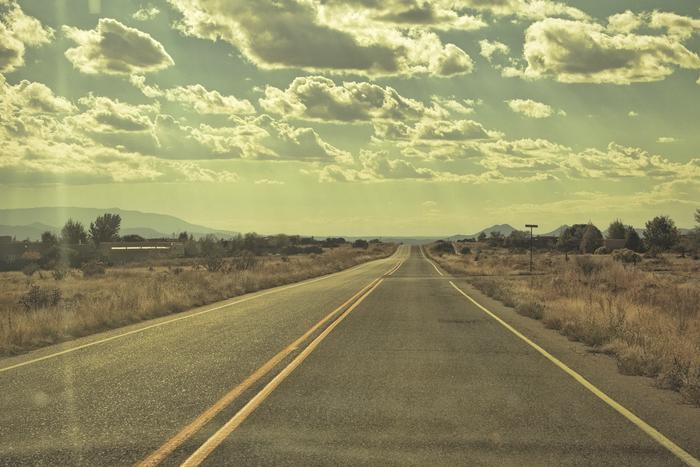}
\end{subfigure}%
\begin{subfigure}{.2\textwidth}
	\includegraphics[width=0.95\linewidth, right]{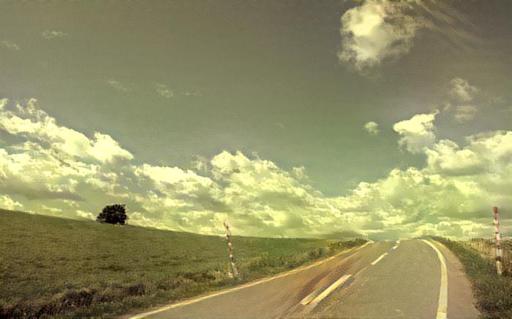}
\end{subfigure}%
\begin{subfigure}{.2\textwidth}
	\includegraphics[width=0.95\linewidth, right]{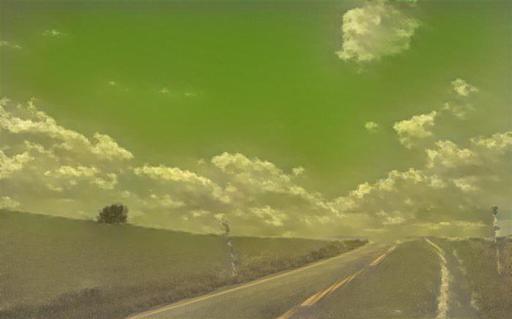}
\end{subfigure}%
\begin{subfigure}{.2\textwidth}
	\includegraphics[width=0.95\linewidth, right]{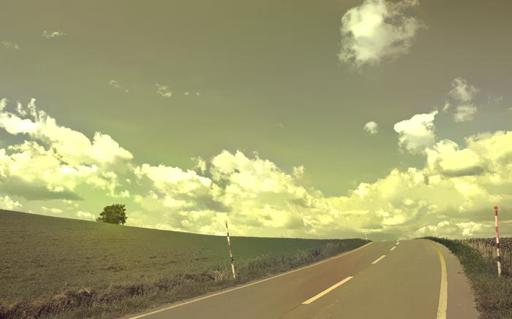}
\end{subfigure}%

\begin{subfigure}{.2\textwidth}
	\includegraphics[width=0.95\linewidth, left]{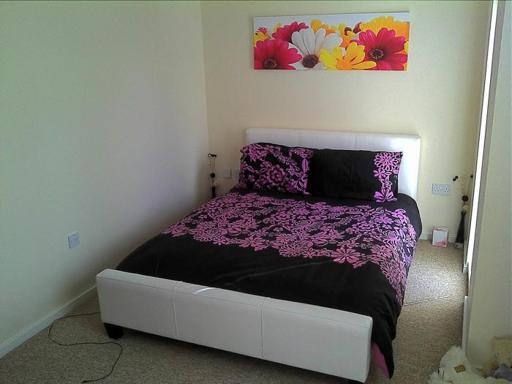}
\end{subfigure}%
\begin{subfigure}{.2\textwidth}
	\includegraphics[width=0.95\linewidth, left]{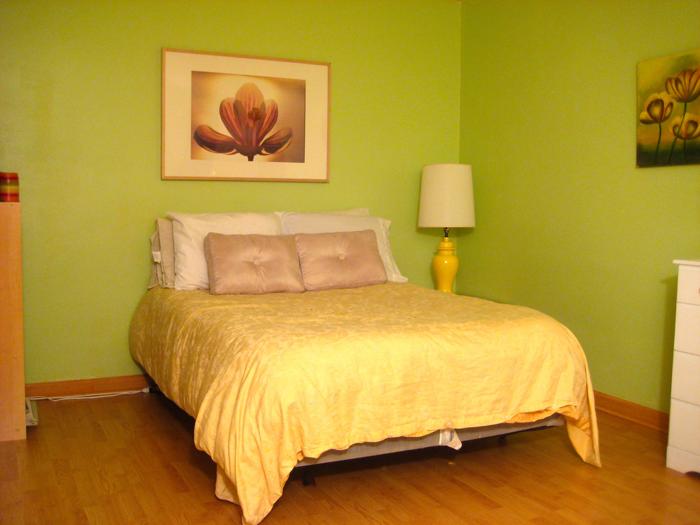}
\end{subfigure}%
\begin{subfigure}{.2\textwidth}
	\includegraphics[width=0.95\linewidth, right]{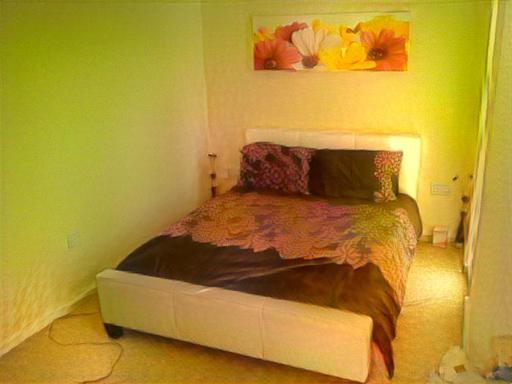}
\end{subfigure}%
\begin{subfigure}{.2\textwidth}
	\includegraphics[width=0.95\linewidth, right]{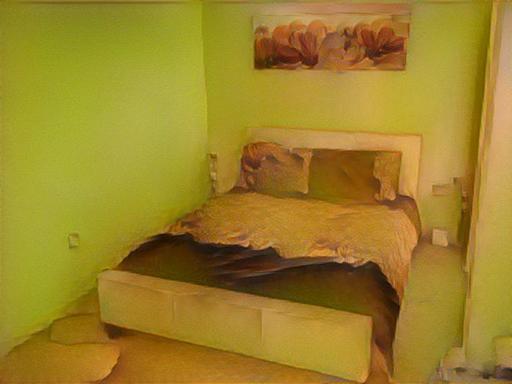}
\end{subfigure}%
\begin{subfigure}{.2\textwidth}
	\includegraphics[width=0.95\linewidth, right]{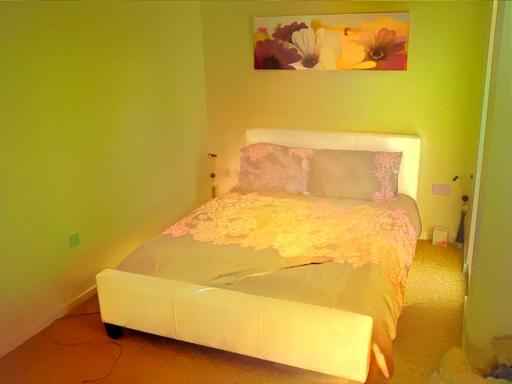}
\end{subfigure}%

\begin{subfigure}{.2\textwidth}
	\includegraphics[width=0.95\linewidth, left]{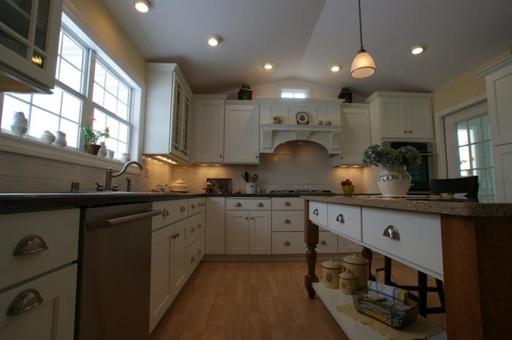}
	\caption{Input image}
\end{subfigure}%
\begin{subfigure}{.2\textwidth}
	\includegraphics[width=0.95\linewidth, left]{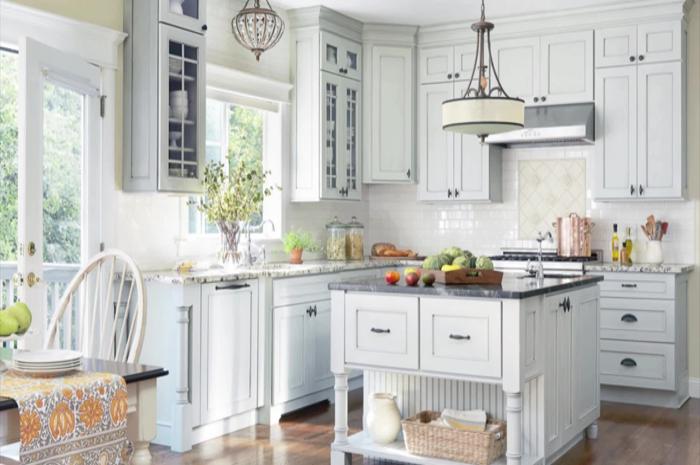}
	\caption{Reference style image}
\end{subfigure}%
\begin{subfigure}{.2\textwidth}
	\includegraphics[width=0.95\linewidth, right]{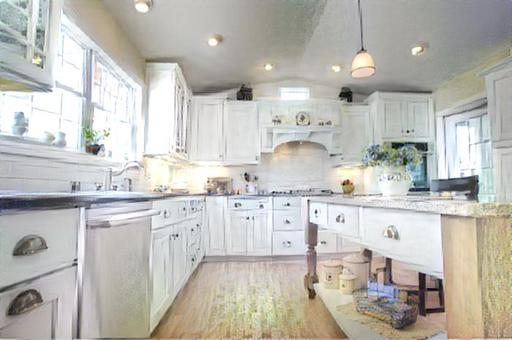}
	\caption{Neural Style}
\end{subfigure}%
\begin{subfigure}{.2\textwidth}
	\includegraphics[width=0.95\linewidth, right]{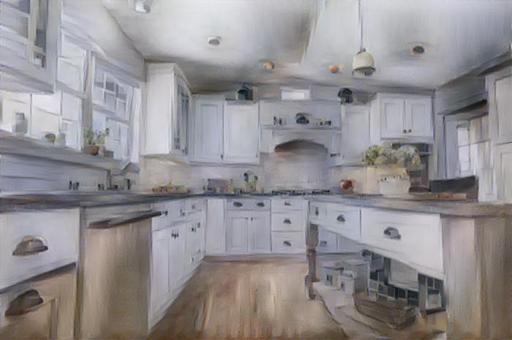}
	\caption{CNNMRF }
\end{subfigure}%
\begin{subfigure}{.2\textwidth}
	\includegraphics[width=0.95\linewidth, right]{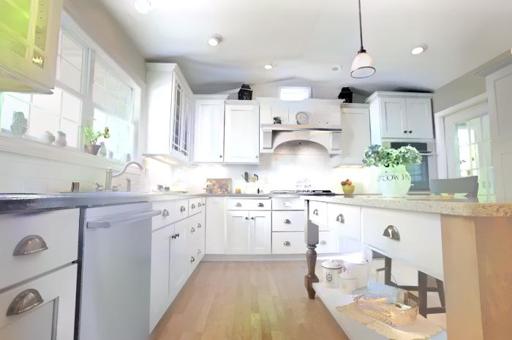}
	\caption{Our result}
\end{subfigure}%
 
\caption{Comparison of our method against Neural Style and CNNMRF. Both Neural Style and CNNMRF produce strong distortions in their synthesized images. Neural Style also entirely ignores the semantic context for style transfer. CNNMRF tends to ignore most of the texture in the reference style image since it uses nearest neighbor search.  Our approach is free of distortions and matches texture semantically.  }
 
\label{fig:comp1}
\end{figure*}

\begin{figure*}[htp]
\begin{subfigure}{.2\textwidth}
	\includegraphics[width=0.95\linewidth, left]{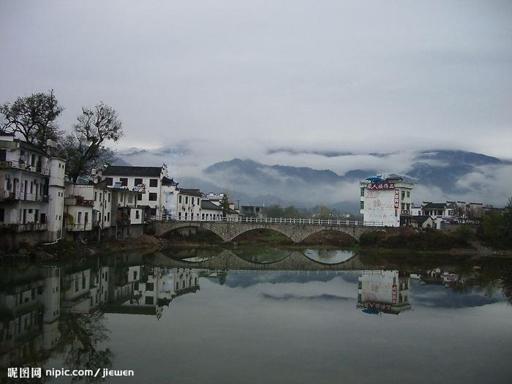}
\end{subfigure}%
\begin{subfigure}{.2\textwidth}
	\includegraphics[width=0.95\linewidth, left]{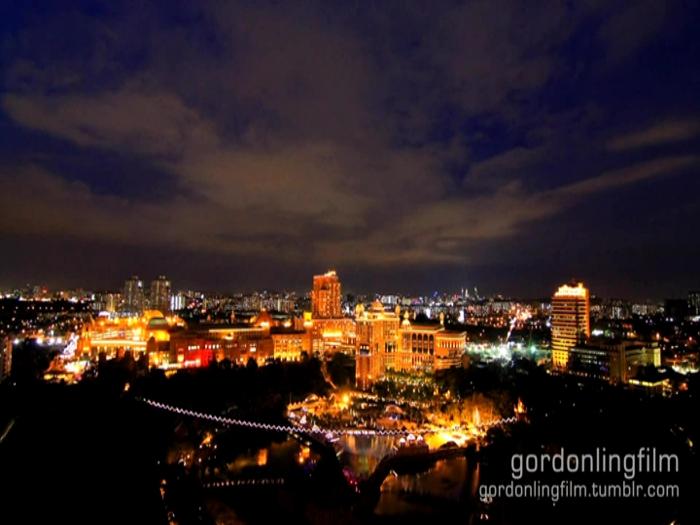}
\end{subfigure}%
\begin{subfigure}{.2\textwidth}
	\includegraphics[width=0.95\linewidth, right]{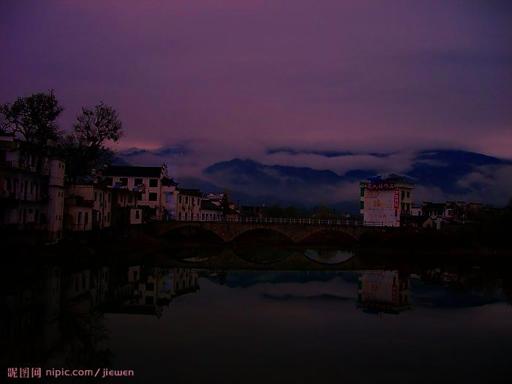}
\end{subfigure}%
\begin{subfigure}{.2\textwidth}
	\includegraphics[width=0.95\linewidth, right]{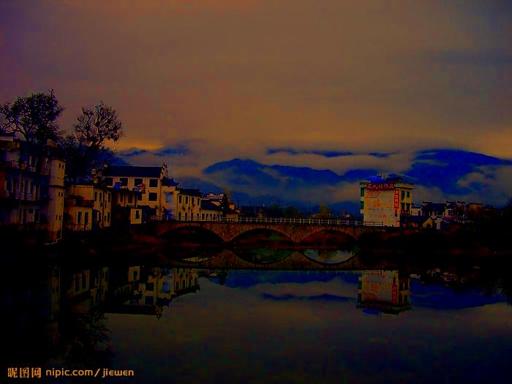}
\end{subfigure}%
\begin{subfigure}{.2\textwidth}
	\includegraphics[width=0.95\linewidth, right]{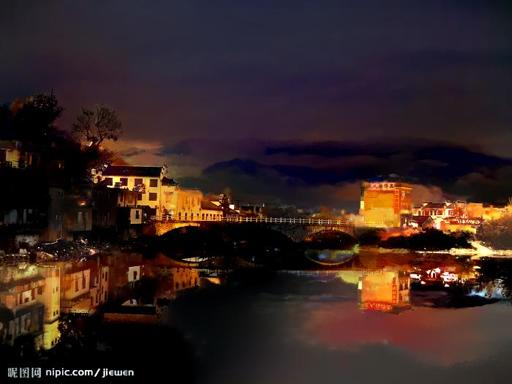}
\end{subfigure}%

\begin{subfigure}{.2\textwidth}
	\includegraphics[width=0.95\linewidth, left]{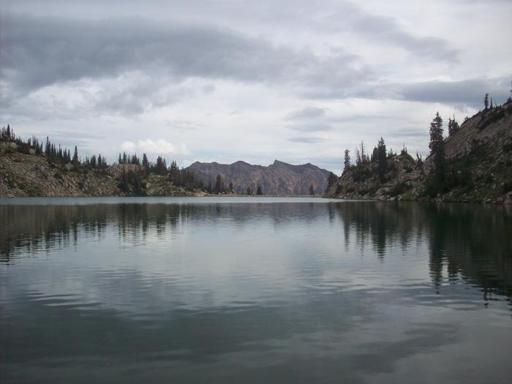}
\end{subfigure}%
\begin{subfigure}{.2\textwidth}
	\includegraphics[width=0.95\linewidth, left]{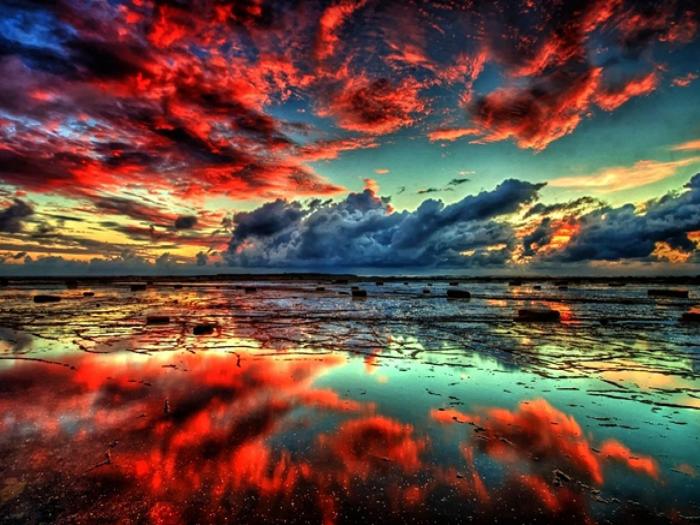}
\end{subfigure}%
\begin{subfigure}{.2\textwidth}
	\includegraphics[width=0.95\linewidth, right]{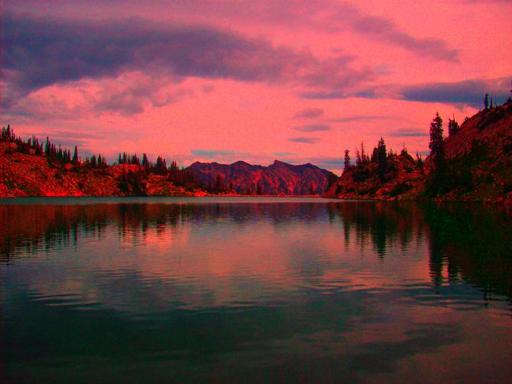}
\end{subfigure}%
\begin{subfigure}{.2\textwidth}
	\includegraphics[width=0.95\linewidth, right]{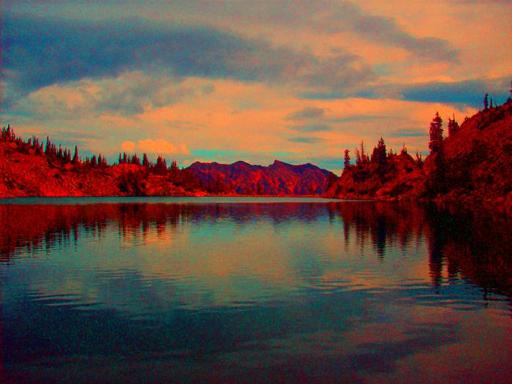}
\end{subfigure}%
\begin{subfigure}{.2\textwidth}
	\includegraphics[width=0.95\linewidth, right]{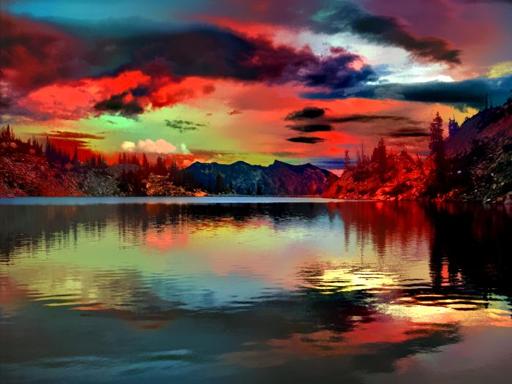}
\end{subfigure}%

\begin{subfigure}{.2\textwidth}
	\includegraphics[width=0.95\linewidth, left]{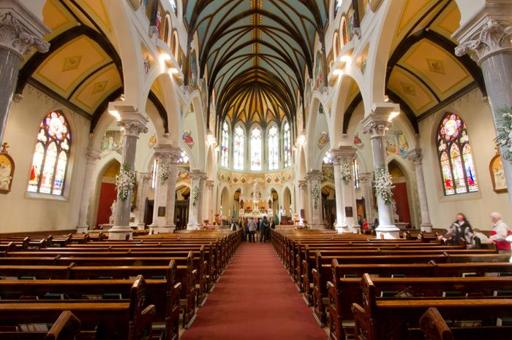}
\end{subfigure}%
\begin{subfigure}{.2\textwidth}
	\includegraphics[width=0.95\linewidth, left]{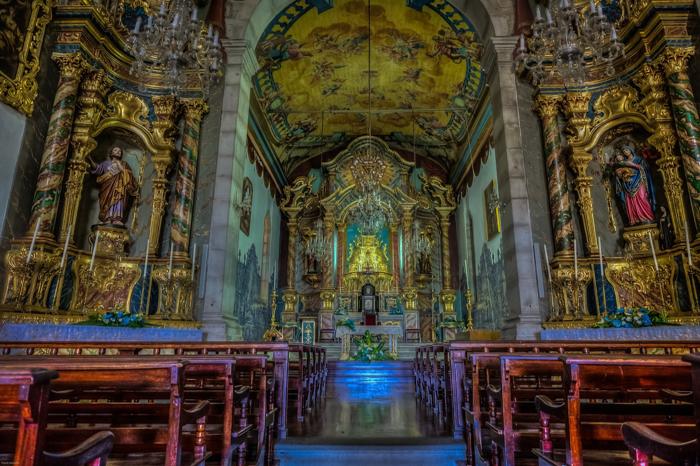}
\end{subfigure}%
\begin{subfigure}{.2\textwidth}
	\includegraphics[width=0.95\linewidth, right]{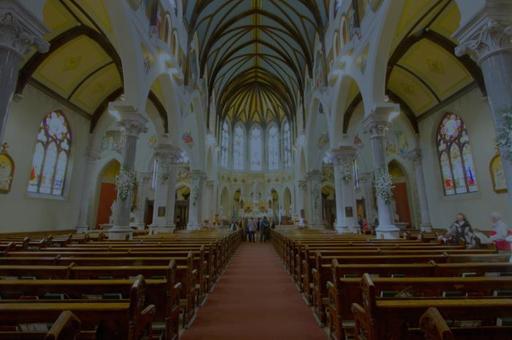}
\end{subfigure}%
\begin{subfigure}{.2\textwidth}
	\includegraphics[width=0.95\linewidth, right]{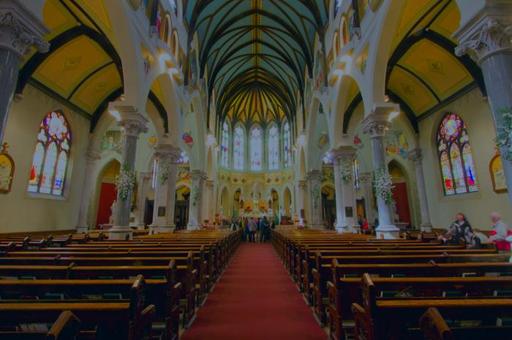}
\end{subfigure}%
\begin{subfigure}{.2\textwidth}
	\includegraphics[width=0.95\linewidth, right]{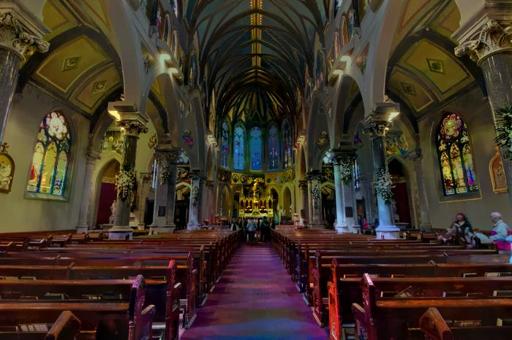}
\end{subfigure}%

\begin{subfigure}{.2\textwidth}
	\includegraphics[width=0.95\linewidth, left]{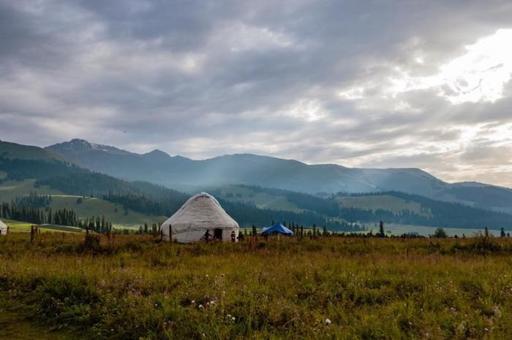}
\end{subfigure}%
\begin{subfigure}{.2\textwidth}
	\includegraphics[width=0.95\linewidth, left]{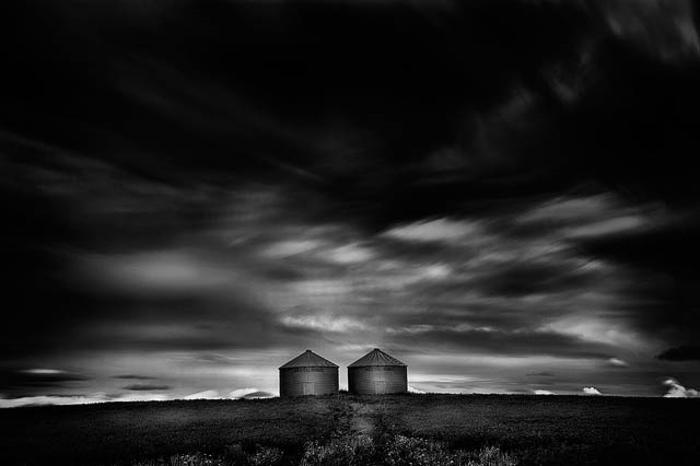}
\end{subfigure}%
\begin{subfigure}{.2\textwidth}
	\includegraphics[width=0.95\linewidth, right]{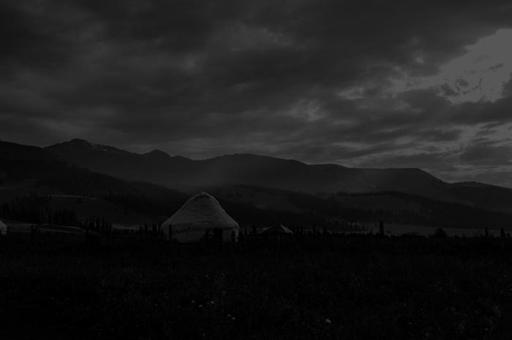}
\end{subfigure}%
\begin{subfigure}{.2\textwidth}
	\includegraphics[width=0.95\linewidth, right]{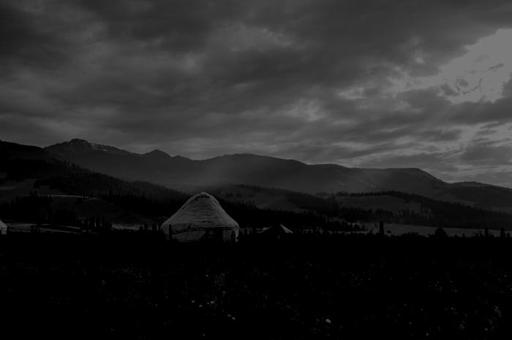}
\end{subfigure}%
\begin{subfigure}{.2\textwidth}
	\includegraphics[width=0.95\linewidth, right]{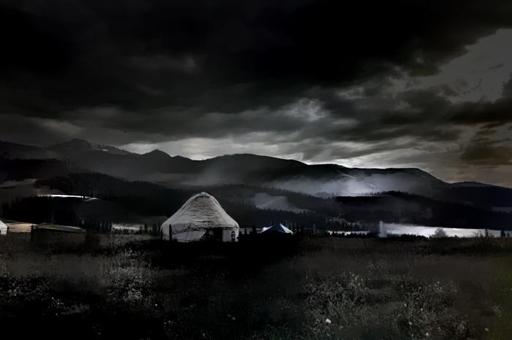}
\end{subfigure}%

\begin{subfigure}{.2\textwidth}
	\includegraphics[width=0.95\linewidth, left]{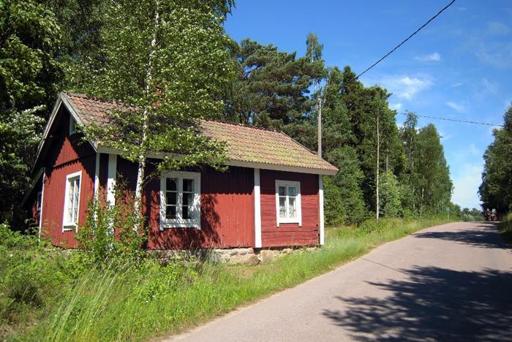}
\end{subfigure}%
\begin{subfigure}{.2\textwidth}
	\includegraphics[width=0.95\linewidth, left]{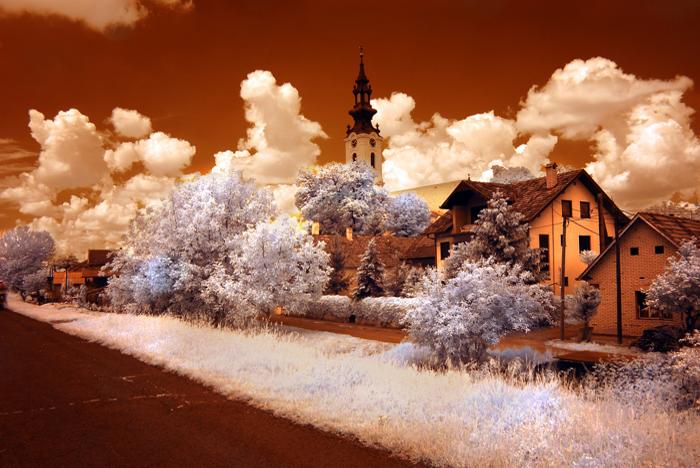}
\end{subfigure}%
\begin{subfigure}{.2\textwidth}
	\includegraphics[width=0.95\linewidth, right]{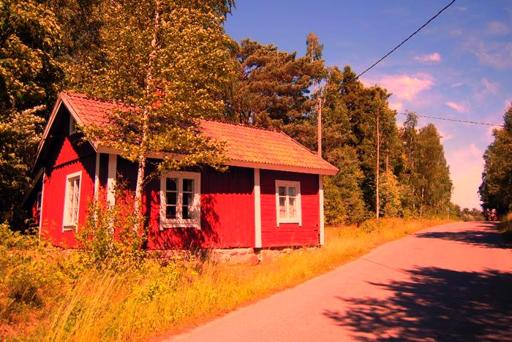}
\end{subfigure}%
\begin{subfigure}{.2\textwidth}
	\includegraphics[width=0.95\linewidth, right]{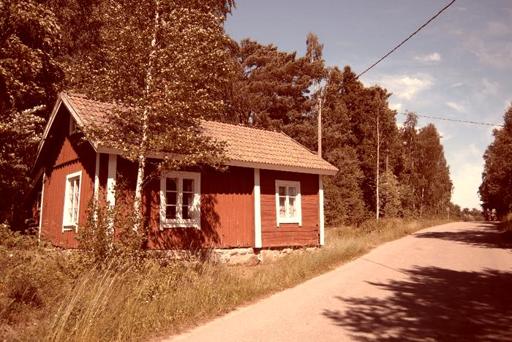}
\end{subfigure}%
\begin{subfigure}{.2\textwidth}
	\includegraphics[width=0.95\linewidth, right]{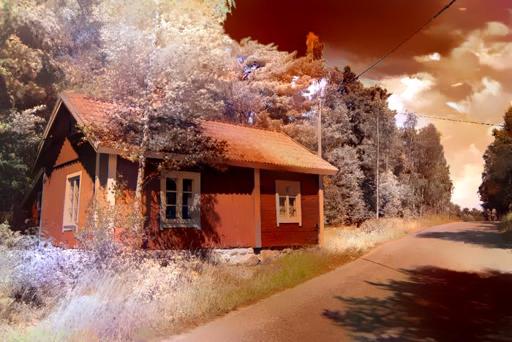}
\end{subfigure}%

\begin{subfigure}{.2\textwidth}
	\includegraphics[width=0.95\linewidth, left]{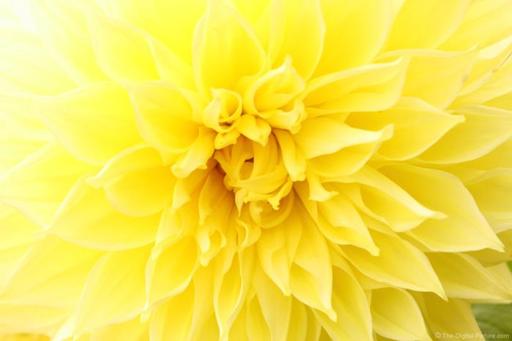}
\end{subfigure}%
\begin{subfigure}{.2\textwidth}
	\includegraphics[width=0.95\linewidth, left]{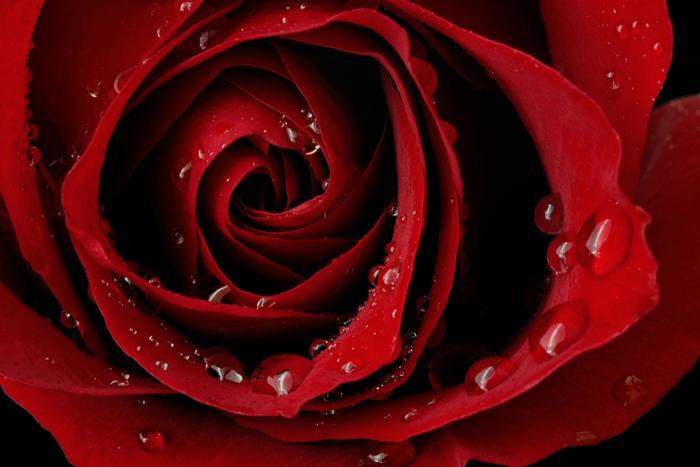}
\end{subfigure}%
\begin{subfigure}{.2\textwidth}
	\includegraphics[width=0.95\linewidth, right]{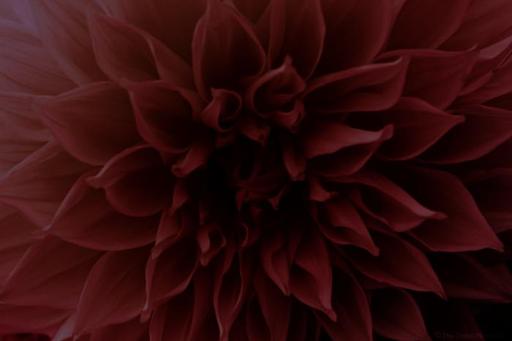}
\end{subfigure}%
\begin{subfigure}{.2\textwidth}
	\includegraphics[width=0.95\linewidth, right]{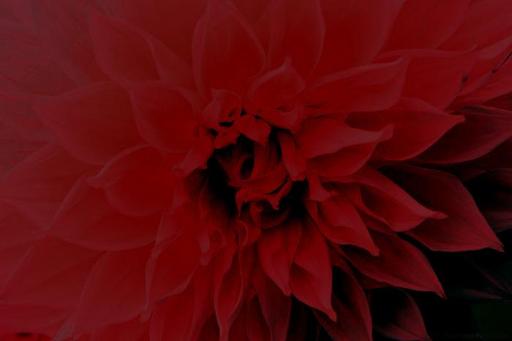}
\end{subfigure}%
\begin{subfigure}{.2\textwidth}
	\includegraphics[width=0.95\linewidth, right]{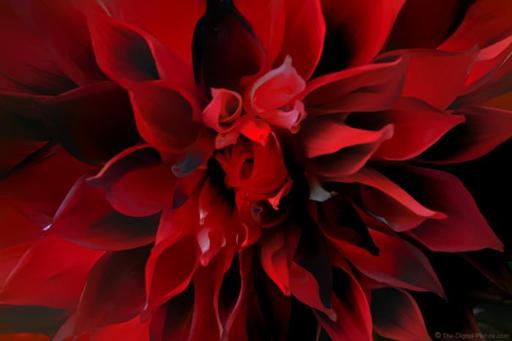}
\end{subfigure}%

\begin{subfigure}{.2\textwidth}
	\includegraphics[width=0.95\linewidth, left]{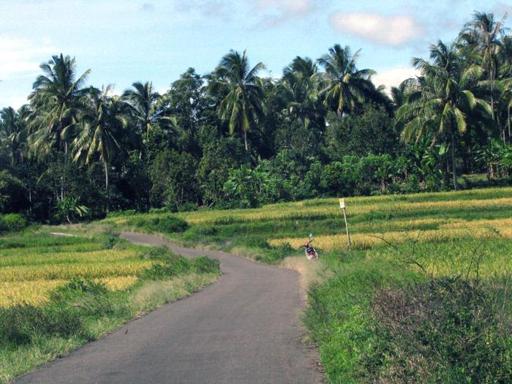}
\end{subfigure}%
\begin{subfigure}{.2\textwidth}
	\includegraphics[width=0.95\linewidth, left]{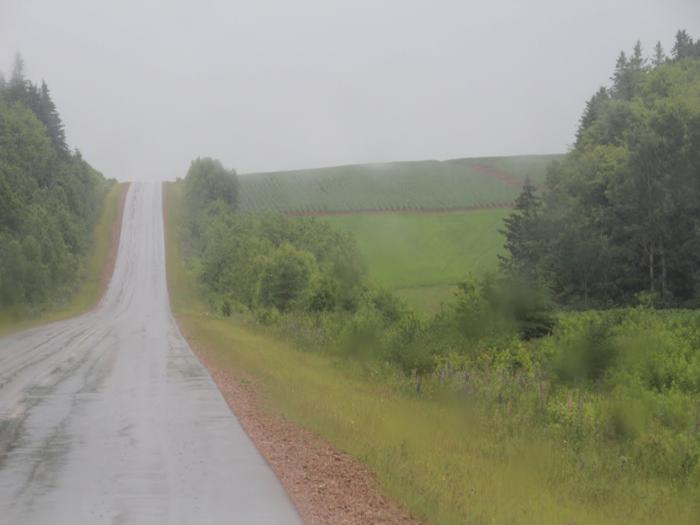}
\end{subfigure}%
\begin{subfigure}{.2\textwidth}
	\includegraphics[width=0.95\linewidth, right]{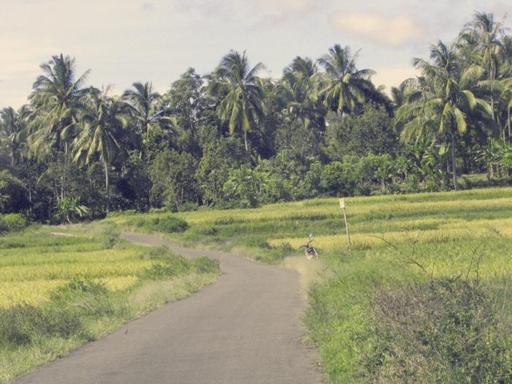}
\end{subfigure}%
\begin{subfigure}{.2\textwidth}
	\includegraphics[width=0.95\linewidth, right]{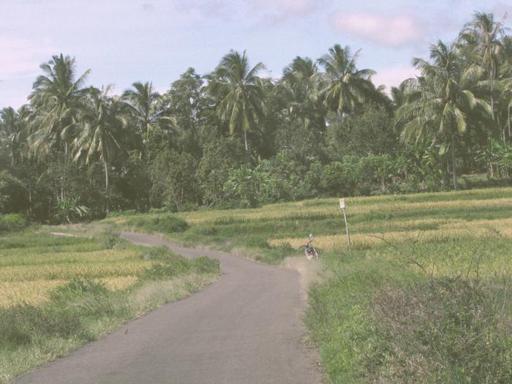}
\end{subfigure}%
\begin{subfigure}{.2\textwidth}
	\includegraphics[width=0.95\linewidth, right]{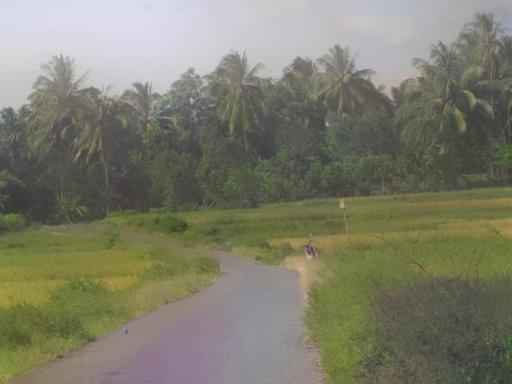}
\end{subfigure}%

\begin{subfigure}{.2\textwidth}
	\includegraphics[width=0.95\linewidth, left]{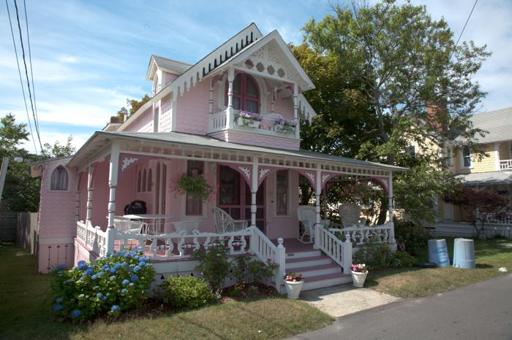}
	\caption{Input image}
\end{subfigure}%
\begin{subfigure}{.2\textwidth}
	\includegraphics[width=0.95\linewidth, left]{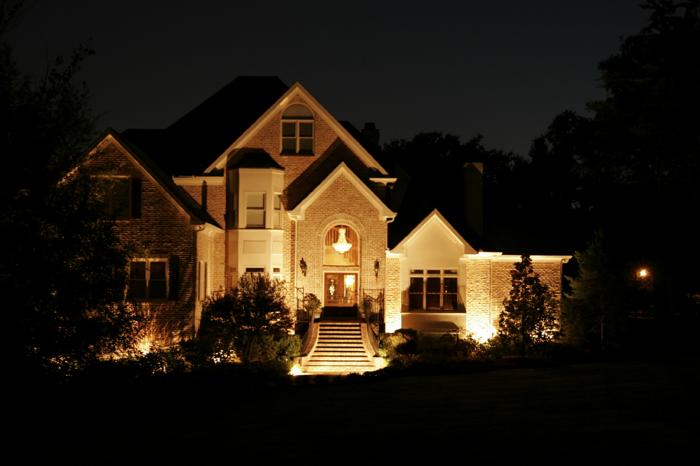}
	\caption{Reference style image}
\end{subfigure}%
\begin{subfigure}{.2\textwidth}
	\includegraphics[width=0.95\linewidth, right]{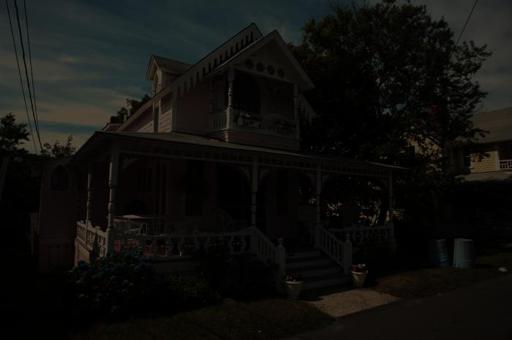}
	\caption{Reinhard et al. \cite{reinhard2001color}}
\end{subfigure}%
\begin{subfigure}{.2\textwidth}
	\includegraphics[width=0.95\linewidth, right]{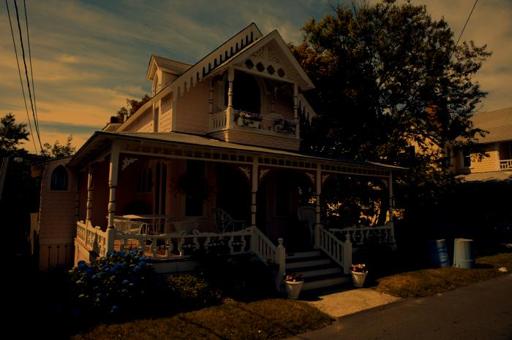}
	\caption{Piti\'e et al. \cite{pitie2005n}}
\end{subfigure}%
\begin{subfigure}{.2\textwidth}
	\includegraphics[width=0.95\linewidth, right]{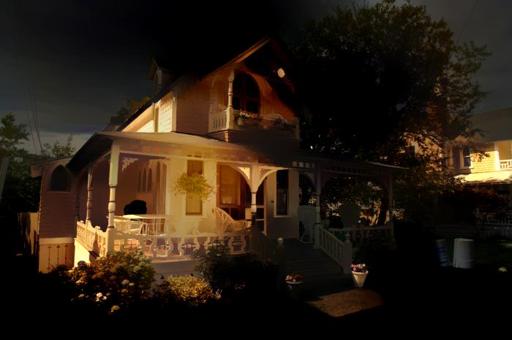}
	\caption{Our result}
\end{subfigure}%
\caption{Comparison of our method against Reinhard et al. \cite{reinhard2001color} and Piti\'e \cite{pitie2005n}. Our method provides more flexibility in transferring spatially-variant color changes, yielding better results than previous techniques. }
\label{fig:comp2}
\end{figure*}

\section{Implementation Details}

This section describes the implementation details of our approach.  We
employed the pre-trained VGG-19 \cite{simonyan2014very} as the feature
extractor. We chose \convlayer{conv4\_2} ($\alpha_\ell = 1$ for this
layer and $\alpha_\ell=0$ for all other layers) as the content
representation, and \convlayer{conv1\_1}, \convlayer{conv2\_1},
\convlayer{conv3\_1}, \convlayer{conv4\_1} and \convlayer{conv5\_1}
($\beta_\ell=1/5$ for those layers and $\beta_\ell=0$ for all other
layers) as the style representation.  We used these layer preferences
and parameters $\Gamma = 10^2$, $\lambda=10^4$ for all the results. {The effect of $\lambda$ is illustrated in Figure~\ref{fig:lambda}}.

We use the original author's Matlab implementation of Levin et
al.~\cite{levin2008closed} to compute the Matting Laplacian matrices
and modified the publicly available torch
implementation~\cite{Johnson2015} of the Neural Style algorithm. {The derivative of the photorealism regularization term is implemented in CUDA for gradient-based optimization.}

We initialize our optimization using {the output of} the Neural Style algorithm
(Eq.~\ref{eq:neural_style}) with the augmented style loss
(Eq.~\ref{eq:neural_style_s_plus}), which itself is initialized with a
random noise. 
This two-stage optimization works better than solving for
Equation~\ref{eq:ours} directly, as it prevents the suppression of proper
local color transfer due to the strong photorealism regularization.

We use DilatedNet~\cite{chen2016deeplab} for segmenting both the input
image and reference style image. As is, this technique recognizes 150
categories. We found that this fine-grain classification was
unnecessary and a source of instability in our algorithm. We merged
similar classes such as `lake', `river', `ocean', and `water' that are
equivalent in our context to get a reduced set of classes that
yields cleaner and simpler segmentations, and eventually better
outputs. The merged labels are detailed in the supplemental material.

Our code is available at: \url{https://github.com/luanfujun/deep-photo-styletransfer}.

\section{Results and Comparison}

We have performed a series of experiments to validate our approach. We
first discuss visual comparisons with previous work before reporting
the results of two user studies.

We compare our method with Gatys et al.~\cite{gatys2016image}
(\emph{Neural Style} for short) and Li et al.~\cite{li2016combining}
(\emph{CNNMRF} for short) across a series of indoor and outdoor scenes
in Figure~\ref{fig:comp1}. 
Both techniques produce results with painting-like distortions, which are
undesirable in the context of photographic style transfer. The Neural
Style algorithm also suffers from spillovers in several cases, e.g.,
with the sky taking on the style of the ground. And as previously
discussed, CNNMRF often generates partial style transfers that ignore
significant portions of the style image. In comparison, our
photorealism regularization and semantic segmentation prevent these
artifacts from happening and our results look visually more
satisfying.

In Figure~\ref{fig:comp2}, we compare our method with global style
transfer methods that do not distort images, Reinhard et
al.~\cite{reinhard2001color} and Piti\'e et
al.~\cite{pitie2005n}. Both techniques apply a global color mapping to
match the color statistics between the input image and the style
image, which limits the faithfulness of their results when the
transfer requires spatially-varying color transformation. Our transfer
is local and capable of handling context-sensitive color changes.

\begin{figure*}[htp]
\centering

\subcaptionbox{Input image}
  [.25\textwidth]{\includegraphics[width=0.245\linewidth]{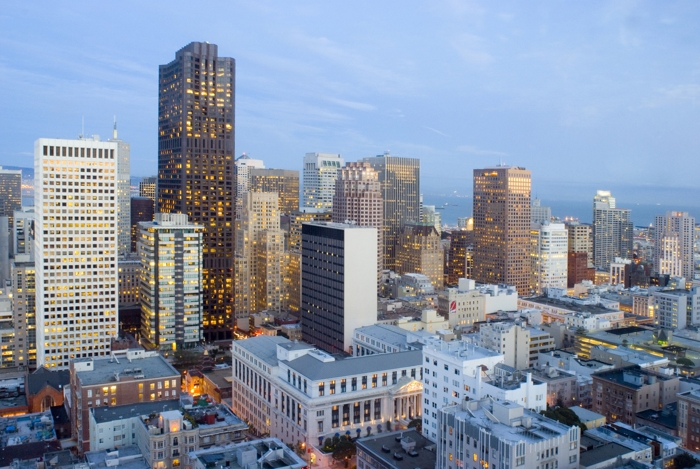}}%
\subcaptionbox{Reference style image}
  [.25\textwidth]{\includegraphics[width=0.245\linewidth]{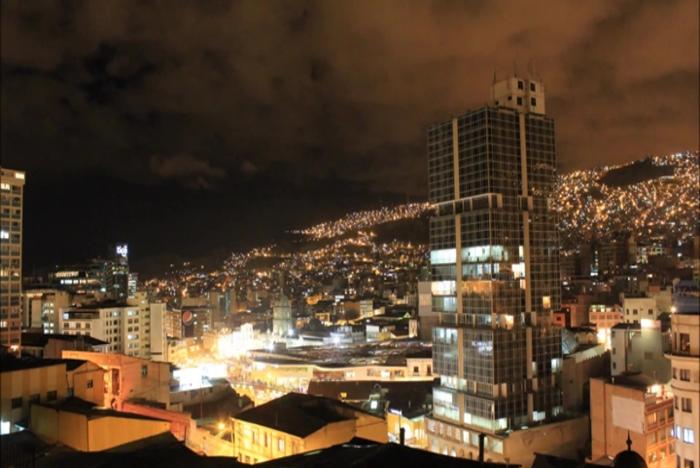}}%
\subcaptionbox{Our result}
  [.25\textwidth]{\includegraphics[width=0.245\linewidth]{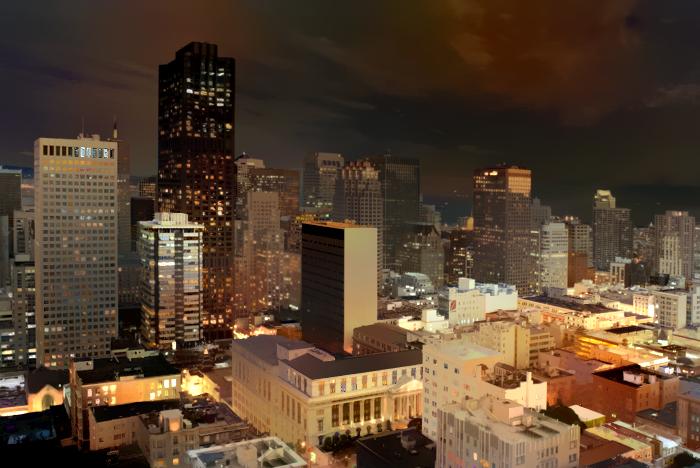}}%
\subcaptionbox{Shih et al. \cite{shih2013data}}
  [.25\textwidth]{\includegraphics[width=0.245\linewidth]{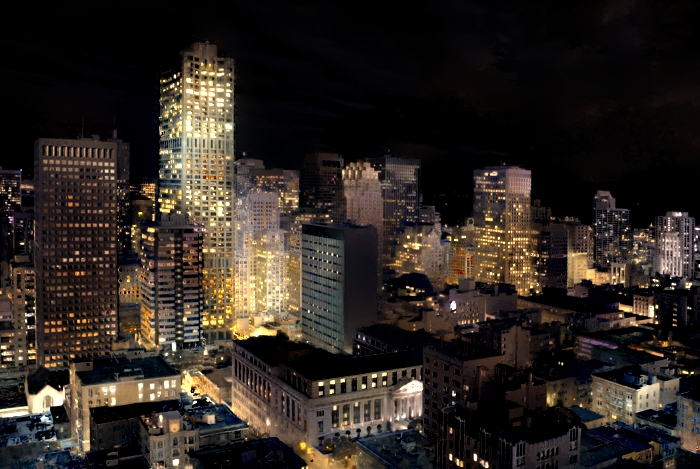}}

	\caption{Our method and the techique of Shih et
          al.~\cite{shih2013data} generate visually satisfying results. However, our algorithm requires
        a single style image instead of a full time-lapse video, and
        it can handle other scenarios in addition to time-of-day hallucination.}
\label{fig:timeofday}
\end{figure*}

In Figure~\ref{fig:timeofday}, we compare our method with the
time-of-day hallucination of Shih et al.~\cite{shih2013data}. The two
results look drastically different because our algorithm directly reproduces
the style of the reference style image whereas Shih's is an
analogy-based technique that transfers the color change observed in a
time-lapse video. Both results are visually satisfying and we believe
that which one is most useful depends on the application. From a technical perspective, our approach is more
practical because it requires only a single style photo in addition to
the input picture whereas Shih's hallucination needs a full time-lapse
video, which is a less common medium and requires 
more storage. Further, our algorithm can handle other scenarios beside
time-of-day hallucination.

\begin{figure}[htp]
\centering

\subcaptionbox{Input}
  [.1666\textwidth]{\includegraphics[width=0.33333\linewidth]{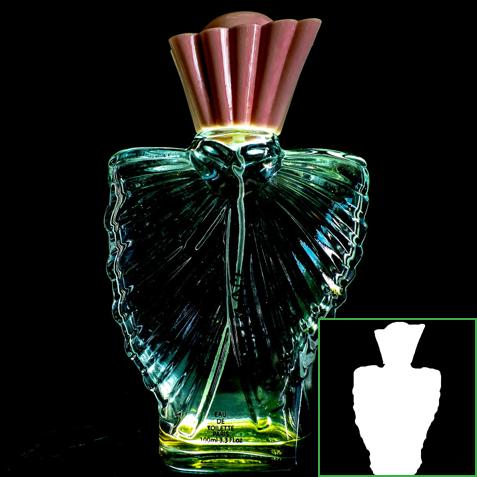}}%
\subcaptionbox{Reference style}
  [.1666\textwidth]{\includegraphics[width=0.33333\linewidth]{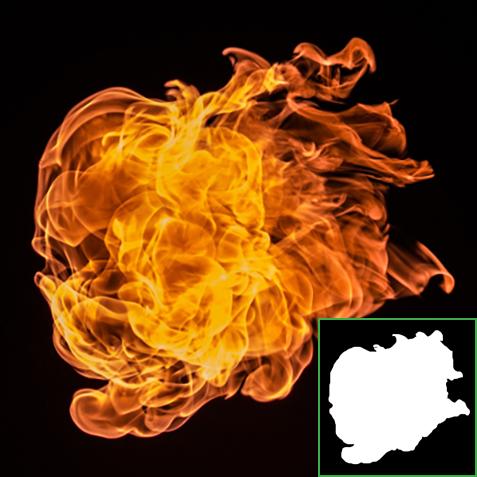}}%
\subcaptionbox{Our result}
  [.1666\textwidth]{\includegraphics[width=0.33333\linewidth]{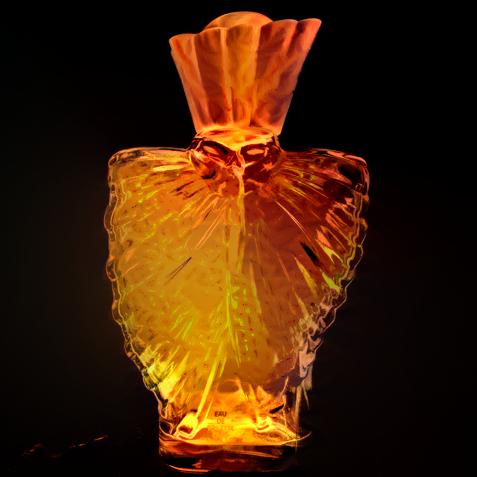}}

\subcaptionbox{Input}
  [.25\textwidth]{\includegraphics[width=0.495\linewidth]{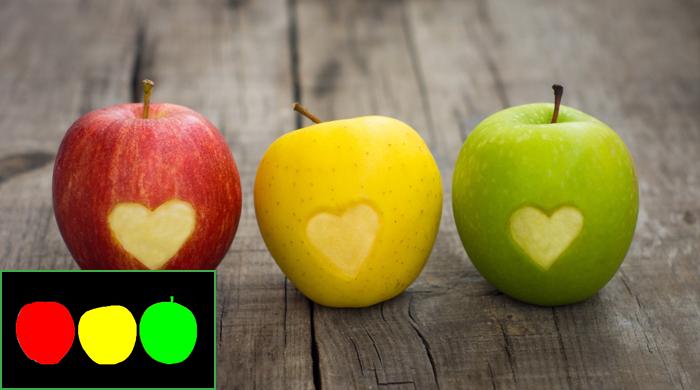}}%
\subcaptionbox{Our result}
  [.25\textwidth]{\includegraphics[width=0.495\linewidth]{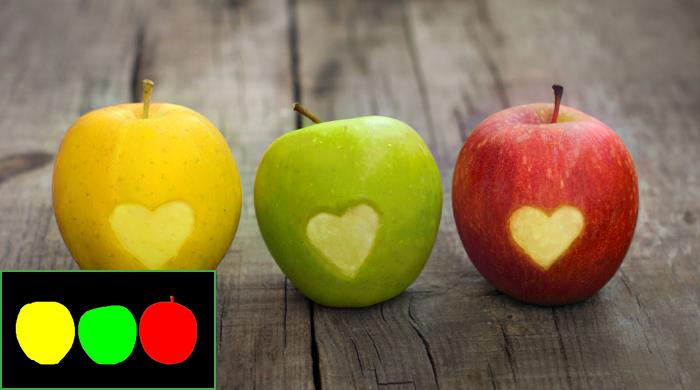}}

\caption{Manual segmentation enables diverse tasks such as
  transferring a fireball~(b) to a perfume bottle~(a) to produce a
  fire-illuminated look~(c), or switching the texture between
  different apples~(d, e). }
\label{fig:editing}
\end{figure}

\begin{figure}[htp]
\centering
\begin{subfigure}{.166\textwidth}
	\includegraphics[width=0.95\linewidth]{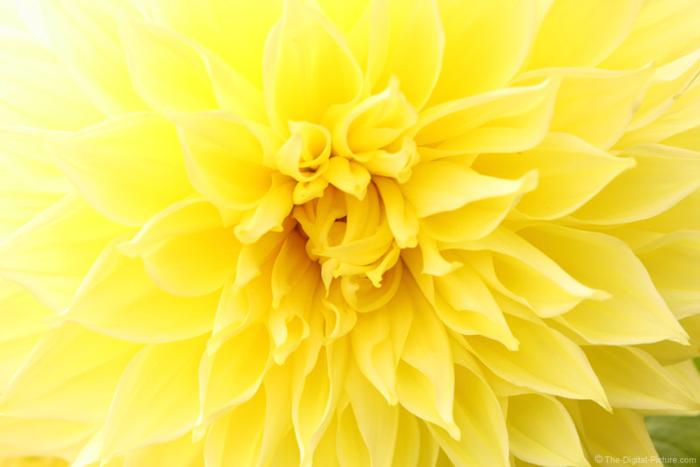}
\end{subfigure}%
\begin{subfigure}{.166\textwidth}
	\includegraphics[width=0.95\linewidth]{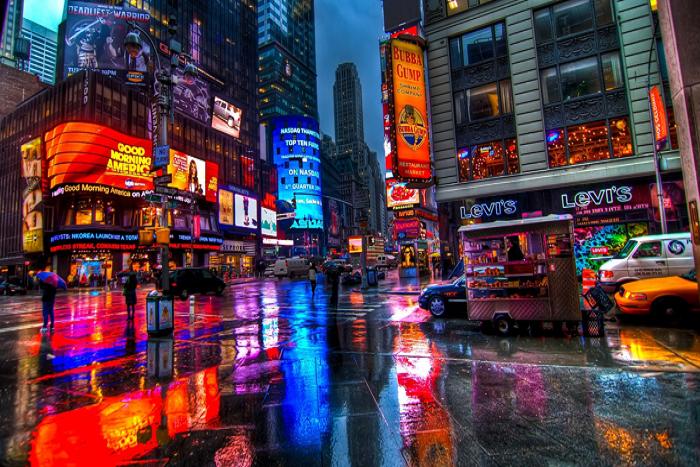}
\end{subfigure}%
\begin{subfigure}{.166\textwidth}
	\includegraphics[width=0.95\linewidth]{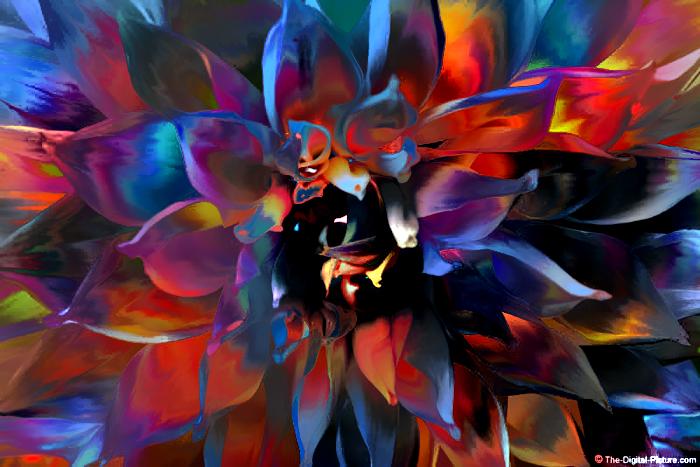}
\end{subfigure}

\begin{subfigure}{.166\textwidth}
	\includegraphics[width=0.95\linewidth]{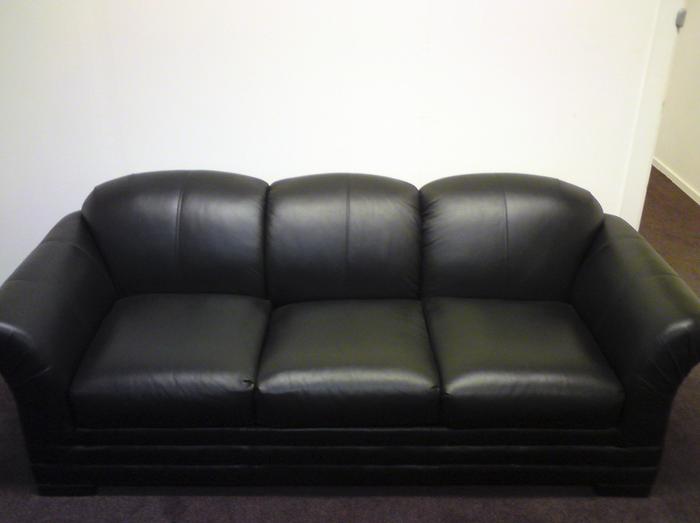}
	\caption{Input}
\end{subfigure}%
\begin{subfigure}{.166\textwidth}
	\includegraphics[width=0.95\linewidth]{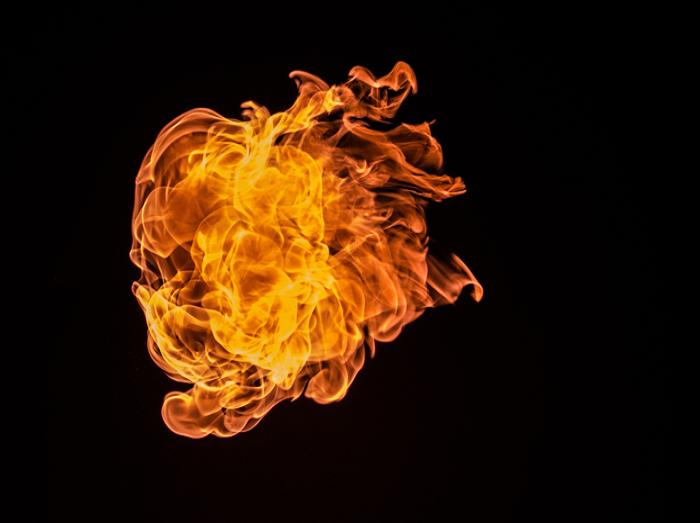}
	\caption{Reference style}
\end{subfigure}%
\begin{subfigure}{.166\textwidth}
	\includegraphics[width=0.95\linewidth]{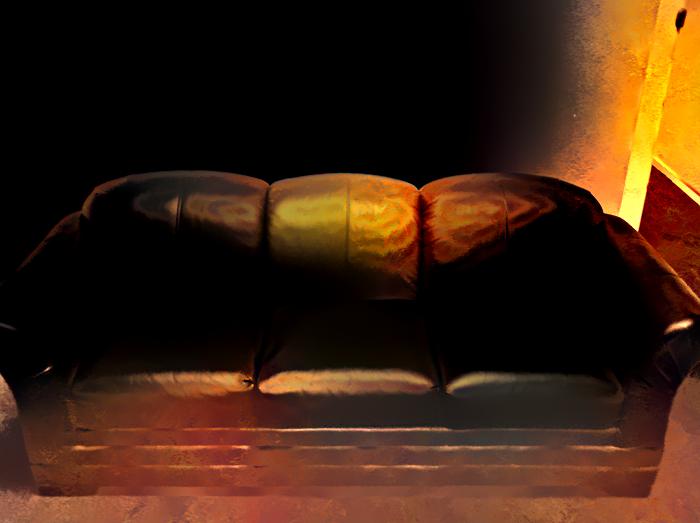}
	\caption{Our result}
\end{subfigure}%

\caption{Failure cases due to extreme mismatch.}
  \label{fig:failure_case}
\end{figure}

In Figure~\ref{fig:editing}, we show how users can control the
transfer results simply by providing the semantic masks. This use case
enables artistic applications and also makes it possible to handle
extreme cases for which semantic labeling cannot help, e.g., to match a transparent perfume bottle to a fireball. 

In Figure~\ref{fig:failure_case}, we show examples of failure due to extreme mismatch. These can be fixed using manual segmentation.

We provide additional results such as comparison against Wu et al.~\cite{wu2013content}, results with only semantic segmentation or photorealism regularization applied separately, and a solution for handling noisy or high-resolution input in the supplemental material. All our results were generated using a two-stage optimization in 3\char`\~5 minutes on an NVIDIA Titan X GPU.

\paragraph{User studies.}

\begin{figure}[htp]
\centering
\includegraphics[width=1.0\linewidth]{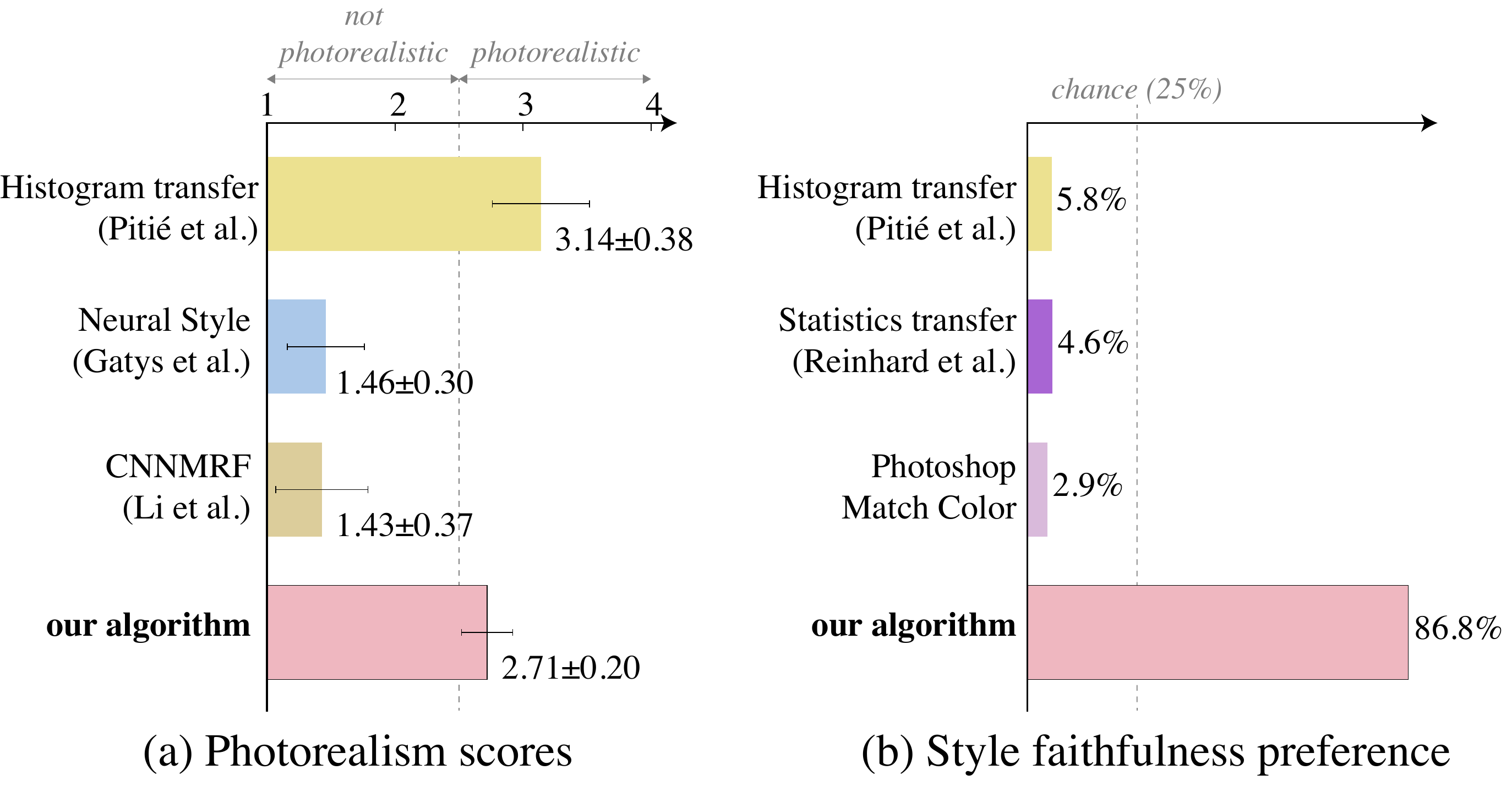}
\caption{User study results confirming that our algorithm produces
  photorealistic and faithful results.}
\label{fig:survey}
\end{figure}

We conducted two user studies to validate our work. First, we assessed
the photorealism of several techniques: ours, the histogram transfer
of Piti\'e et al.~\cite{pitie2005n}, CNNMRF~\cite{li2016combining},
and Neural Style~\cite{gatys2016image}. We asked users to score images
on a 1-to-4 scale ranging from ``definitely not photorealistic'' to
``definitely photorealistic''. We used 8 different scenes for each of
the 4 methods for a total of 32 questions. We collected 40 responses
per question on average. Figure~\ref{fig:survey}a shows that CNNMRF
and Neural Style produce nonphotorealistic results, which confirms our
observation that these techniques introduce painting-like
distortions. It also shows that, although our approach scores below
histogram transfer, it nonetheless produces photorealistic
outputs. Motivated by this result, we conducted a second study to
estimate the faithfulness of the style transfer techniques. We found that
global methods consistently generated distortion-free results but with a
variable level of style faithfulness. We compared against several global
methods in our second study: Reinhard's statistics
transfer~\cite{reinhard2001color}, Piti\'e's histogram
transfer~\cite{pitie2005n}, and Photoshop Match Color.  
Users were shown a style image and 4 transferred outputs, the 3 previously mentioned
global methods and our technique (randomly ordered to avoid bias), and asked
to choose the image with the most similar style to the reference style image.
We, on purpose, did not show the input image so that users could focus
on the output images.  We showed 20 comparisons and collected 35 responses per
question on average. The study shows that our algorithm produces the most
faithful style transfer results more than 80\% of the time (Fig.~\ref{fig:survey}b). \fl{We provide the links to our user study websites in the supplemental material.}

\section{Conclusions}
We introduce a deep-learning approach that faithfully transfers style from a reference image for a wide variety of image content. We use the Matting Laplacian to constrain the transformation from the input to the output to be locally affine in colorspace. Semantic segmentation further drives more meaningful style transfer yielding satisfying photorealistic results in a broad variety of scenarios, including transfer of the time of day, weather, season, and artistic edits.

\section*{Acknowledgments}
We thank Leon Gatys, Fr\'edo Durand, Aaron Hertzmann as well as the anonymous reviewers for their valuable discussions. We thank Fuzhang Wu for generating results using \cite{wu2013content}. This research is supported by a Google Faculty Research Award, and NSF awards IIS 1617861 and 1513967.

{\small
\bibliographystyle{ieee}
\bibliography{egbib}

\begin{thebibliography}{10}\itemsep=-1pt

\bibitem{bae2006two}
S.~Bae, S.~Paris, and F.~Durand.
\newblock Two-scale tone management for photographic look.
\newblock In {\em ACM Transactions on Graphics (TOG)}, volume~25, pages
  637--645. ACM, 2006.

\bibitem{neural_doodle}
A.~J. Champandard.
\newblock Semantic style transfer and turning two-bit doodles into fine
  artworks.
\newblock Mar 2016.

\bibitem{chen2016deeplab}
L.-C. Chen, G.~Papandreou, I.~Kokkinos, K.~Murphy, and A.~L. Yuille.
\newblock Deeplab: Semantic image segmentation with deep convolutional nets,
  atrous convolution, and fully connected crfs.
\newblock {\em arXiv preprint arXiv:1606.00915}, 2016.

\bibitem{gardner15}
J.~R. Gardner, M.~J. Kusner, Y.~Li, P.~Upchurch, K.~Q. Weinberger, K.~Bala, and
  J.~E. Hopcroft.
\newblock Deep manifold traversal: Changing labels with convolutional features.
\newblock {\em CoRR}, abs/1511.06421, 2015.

\bibitem{gatys2016image}
L.~A. Gatys, A.~S. Ecker, and M.~Bethge.
\newblock Image style transfer using convolutional neural networks.
\newblock In {\em Proceedings of the IEEE Conference on Computer Vision and
  Pattern Recognition}, pages 2414--2423, 2016.

\bibitem{hertzmann2001image}
A.~Hertzmann, C.~E. Jacobs, N.~Oliver, B.~Curless, and D.~H. Salesin.
\newblock Image analogies.
\newblock In {\em Proceedings of the 28th annual conference on Computer
  graphics and interactive techniques}, pages 327--340. ACM, 2001.

\bibitem{Johnson2015}
J.~Johnson.
\newblock neural-style.
\newblock \url{https://github.com/jcjohnson/neural-style}, 2015.

\bibitem{laffont2014}
P.-Y. Laffont, Z.~Ren, X.~Tao, C.~Qian, and J.~Hays.
\newblock Transient attributes for high-level understanding and editing of
  outdoor scenes.
\newblock {\em ACM Transactions on Graphics}, 33(4), 2014.

\bibitem{levin2008closed}
A.~Levin, D.~Lischinski, and Y.~Weiss.
\newblock A closed-form solution to natural image matting.
\newblock {\em IEEE Transactions on Pattern Analysis and Machine Intelligence},
  30(2):228--242, 2008.

\bibitem{li2016combining}
C.~Li and M.~Wand.
\newblock Combining markov random fields and convolutional neural networks for
  image synthesis.
\newblock {\em arXiv preprint arXiv:1601.04589}, 2016.

\bibitem{pitie2005n}
F.~Pitie, A.~C. Kokaram, and R.~Dahyot.
\newblock N-dimensional probability density function transfer and its
  application to color transfer.
\newblock In {\em Tenth IEEE International Conference on Computer Vision
  (ICCV'05) Volume 1}, volume~2, pages 1434--1439. IEEE, 2005.

\bibitem{reinhard2001color}
E.~Reinhard, M.~Adhikhmin, B.~Gooch, and P.~Shirley.
\newblock Color transfer between images.
\newblock {\em IEEE Computer Graphics and Applications}, 21(5):34--41, 2001.

\bibitem{selim2016painting}
A.~Selim, M.~Elgharib, and L.~Doyle.
\newblock Painting style transfer for head portraits using convolutional neural
  networks.
\newblock {\em ACM Transactions on Graphics (TOG)}, 35(4):129, 2016.

\bibitem{shih2014style}
Y.~Shih, S.~Paris, C.~Barnes, W.~T. Freeman, and F.~Durand.
\newblock Style transfer for headshot portraits.
\newblock 2014.

\bibitem{shih2013data}
Y.~Shih, S.~Paris, F.~Durand, and W.~T. Freeman.
\newblock Data-driven hallucination of different times of day from a single
  outdoor photo.
\newblock {\em ACM Transactions on Graphics (TOG)}, 32(6):200, 2013.

\bibitem{simonyan2014very}
K.~Simonyan and A.~Zisserman.
\newblock Very deep convolutional networks for large-scale image recognition.
\newblock {\em arXiv preprint arXiv:1409.1556}, 2014.

\bibitem{sunkavalli2010multi}
K.~Sunkavalli, M.~K. Johnson, W.~Matusik, and H.~Pfister.
\newblock Multi-scale image harmonization.
\newblock {\em ACM Transactions on Graphics (TOG)}, 29(4):125, 2010.

\bibitem{gram}
E.~W. Weisstein.
\newblock Gram matrix.
\newblock MathWorld--A Wolfram Web Resource.
\newblock http://mathworld.wolfram.com/GramMatrix.html.

\bibitem{wu2013content}
F.~Wu, W.~Dong, Y.~Kong, X.~Mei, J.-C. Paul, and X.~Zhang.
\newblock Content-based colour transfer.
\newblock In {\em Computer Graphics Forum}, volume~32, pages 190--203. Wiley
  Online Library, 2013.

\end{thebibliography}
}

\end{document}